\documentclass{article} % For LaTeX2e
\usepackage{iclr2027_conference,times}

\usepackage{amsmath,amsfonts,bm}

\def\eqref#1{equation~\ref{#1}}
\def\1{\bm{1}}

\DeclareMathAlphabet{\mathsfit}{\encodingdefault}{\sfdefault}{m}{sl}
\SetMathAlphabet{\mathsfit}{bold}{\encodingdefault}{\sfdefault}{bx}{n}

\usepackage{hyperref}
\usepackage{url}
\usepackage[T1]{fontenc}    % use 8-bit T1 fonts
\usepackage{booktabs}       % professional-quality tables
\usepackage{amsfonts}       % blackboard math symbols
\usepackage{nicefrac}       % compact symbols for 1/2, etc.
\usepackage{microtype}      % microtypography
\usepackage{xcolor}         % colors

\usepackage{subcaption}

\usepackage{graphicx}
\usepackage{amsmath}
\usepackage{cleveref}
\usepackage[most]{tcolorbox}
\usepackage{enumitem}
\usepackage{listings}
\usepackage{colortbl}
\usepackage{algorithm}
\usepackage{algpseudocode}
\usepackage{tabularx}
\usepackage{multirow}

\usepackage{pifont}
\usepackage{arydshln}

\usepackage{tikz}
\usetikzlibrary{calc}

\definecolor{lightblue}{RGB}{245,245,255}

\definecolor{lightgray}{gray}{0.95}
\definecolor{nativeblue}{RGB}{214,234,248}
\definecolor{toolgreen}{RGB}{220,247,220}

\definecolor{bestcolor}{HTML}{E8F5E9}   % light green for best
\definecolor{secondcolor}{HTML}{FFF8E1} % light yellow for second best
\definecolor{grouphead}{HTML}{F5F5F5}   % light gray for group headers
\definecolor{ourscolor}{HTML}{E3F2FD}   % light blue for our method row

\definecolor{poscolor}{HTML}{2E7D32}   % green for positive delta
\definecolor{negcolor}{HTML}{C62828}   % red for negative delta

\newcommand{\yes}{\ding{51}}           % checkmark
\newcommand{\no}{\ding{55}}            % cross

\tcbset{
  chatbox/.style={
    enhanced,
    boxrule=0.5pt,
    sharp corners,
    colframe=black!40,
    fonttitle=\bfseries\small,
    coltitle=black,
    colback=#1,
    left=4pt,
    right=4pt,
    top=2pt,
    bottom=2pt,
    boxsep=2pt,
  }
}

\definecolor{codegray}{gray}{0.95}

\lstdefinestyle{pythonstyle}{
    backgroundcolor=\color{codegray},
    language=Python,
    basicstyle=\ttfamily\small,
    numbers=none,
    breaklines=true,
    showstringspaces=false,
    frame=single,
    framerule=0pt,
    xleftmargin=1em,
    xrightmargin=1em
}

\newtcolorbox{promptbox}[1]{%
  colback=lightblue, colframe=black, arc=4pt, boxrule=1pt,
  enhanced, breakable,
  title={\textbf{#1}}, fonttitle=\bfseries,
  left=4pt, right=4pt, top=4pt, bottom=4pt,
}

\newtcbox{\mytbox}[1][]{colback=gray!5, colframe=black, arc=2pt, boxrule=0.5pt, top=2pt, bottom=2pt, left=2pt, right=2pt, nobeforeafter, enhanced, sharp corners, #1}

\title{GazeVLM: Active Vision via Internal Attention Control for Multimodal Reasoning}

\iclrfinalcopy

\author{%
  Brown Ebouky \\
  IBM Research \\
ETH Zurich \\
  \And
  Gabriele Carrino \\
IBM Research \\
TU Wien \\
  \And
  Niccolo Avogaro \\
IBM Research \\
ETH Zurich \\
  \And
  Christoph Studer \\
ETH Zurich \\
  \And
  Andrea Bartezzaghi \\
IBM Research \\
  \And
  Mattia Rigotti \\
IBM Research \\
}

\begin{document}

\maketitle

\lhead{Preprint.}

\begin{abstract}
Human visual reasoning is governed by active vision, a process where meta-cognitive control drives top-down goal-directed attention, dynamically routing foveal focus toward task-relevant details while maintaining peripheral awareness of the global scene.
% In contrast, modern Vision-Language Models (VLMs) process visual information passively, relying on the static accumulation of massive token contexts that dilute spatial reasoning and induce linguistic hallucinations.
In contrast, modern Vision-Language Models (VLMs) process visual information passively, relying on the static accumulation of massive token contexts that dilute the visual evidence as the reasoning chain grows.
Here we propose GazeVLM, a multimodal model that learns to exert this meta-cognitive oversight over its attention resources by generating gaze actions in its reasoning chain, in the form of \texttt{<LOOK>} tags specifying the coordinates of the region to inspect.
During training, each \texttt{<LOOK>} block triggers a continuous suppression bias on the attention logits that dampens features outside the regions gazed at so far, and that lifts at the end of the block, restoring the global view.
The model is trained with this suppression bias as a teaching signal, first by Supervised Fine-Tuning on curated gaze-reasoning traces, then through Group Relative Policy Optimization (GRPO) with rewards for correct answers and valid grounding.
At deployment, the external suppression is removed and the model, running its unmodified forward pass, steers its own attention to the regions indicated by the \texttt{<LOOK>} blocks, retaining most of the effect of the bias and emulating top-down control of spatial attention.
This model thus transitions between global spatial awareness and localized focal reasoning without relying on external agentic tools like cropping, and without adding visual tokens from re-encoded patches.
Applied to 4B-parameter backbones, GazeVLM demonstrates strong high-resolution multimodal reasoning on HRBench-4k and HRBench-8k, surpassing its base models by about 4 points, and outperforming agentic multimodal pipelines built around thinking with images by 5 to 12 points.
\end{abstract}

\section{Introduction}

Despite the rapid evolution of Vision-Language Models (VLMs) and the successful adaptation of Chain-of-Thought (CoT) prompting to the multimodal domain~\citep{zhang2023multimodalcot}, multi-step visual reasoning remains bottlenecked by a fundamental imbalance between textual generation and visual perception.
Modern VLMs process visual information passively; they rely on the static accumulation of massive token contexts.
As VLMs engage in extended, autoregressive reasoning, this influx of textual data increasingly dominates the causal attention mechanisms, and the original visual features tend to be progressively underweighted.
Consequently, the model often over-relies on linguistic priors and frequently hallucinates objects~\citep{rohrbach2018object, biten2022let, li2023evaluating} or logical steps, drifting away from the spatial content of the image. 

Current paradigms attempt to address this imbalance, but they largely inherit the same passive processing of visual information.
One approach, grounded reasoning, forces models to output bounding box coordinates alongside textual claims \citep{peng2023kosmos2, you2024ferret, man2025argus}. While this enforces a superficial textual link to the image, it does not alter how the decoder attends to the pre-encoded visual features during generation.
Another approach augments VLMs with external ``zoom-in'' tools (e.g., V* \citep{wu2024vstar}, Pixel Reasoner \citep{Wangetal2025}, and Ground-R1 \citep{Caoetal2025}). 
These frameworks isolate relevant details by explicitly cropping and re-encoding new image patches.
While this improves local detail extraction, it incurs a substantial computational penalty: as the recent Adaptive-CoF analysis \citep{zhang2025adaptive} shows, these approaches can add nearly a thousand additional visual tokens to the context window during high-resolution reasoning.
This sparse cropping acts as disjointed snapshots carrying no global position in the image, rather than a cohesive scene, which bloats the model's context with re-encoded pixels.

In human cognition, resolving complex visual queries is not a passive process of statically accumulating pixels or ``cropping'' the visual field.
Instead, it relies on the top-down mechanism of active vision \citep{yarbus1967eye, findlay2003active}.
This is fundamentally a meta-cognitive ability requiring observers to continuously maintain peripheral global awareness of the scene, monitor their understanding, and actively plan where to look and what to attend next \citep{rensink2000dynamic}.
They then dynamically route their focus of attention and foveal gaze to specific local regions, extracting granular details while tuning out background noise \citep{corbetta2002control}.
Humans intuitively distinguish between finding a target (global spatial awareness) and reasoning about its specific details (focal fixation).
Our aim is to equip VLMs with this kind of meta-cognitive capability: an iterative, computationally efficient control loop that flexibly transitions between global and local views without reprocessing new image content.

% In this work, we introduce GazeVLM, a novel visual reasoning framework that internalizes human-like active vision directly within the VLM's architecture. Rather than relying on external cropping tools, GazeVLM applies top-down control directly to the causal attention mask of the pre-encoded visual features. Using its global context, the model meta-cognitively assesses its need for information and generates special \texttt{<LOOK>} tokens paired with bounding box coordinates. This triggers a continuous, plateau-shaped suppression bias that dampens the attention weights of irrelevant visual areas, effectively mimicking foveal fixation, while maintaining zero penalty for the targeted region. Once the local extraction is complete, the model generates a \texttt{</LOOK>} token, which lifts the bias and restores access to the full visual context to dictate its next logical step.

In this work, we introduce GazeVLM, a visual reasoning framework that internalizes human-like active vision into the weights of a standard VLM. Rather than relying on external cropping tools, GazeVLM learns top-down control over how the decoder attends to the pre-encoded visual features. Using its global context, the model meta-cognitively assesses its need for information and generates special \texttt{<LOOK>} tokens paired with bounding box coordinates. During training, each \texttt{<LOOK>} token triggers a continuous, plateau-shaped suppression bias on the attention logits that dampens irrelevant visual areas, emulating covert spatial attention, while maintaining zero penalty for the targeted region. The closing \texttt{</LOOK>} token lifts the bias and restores access to the full visual context to dictate the next logical step.
Crucially, this bias is a teaching signal rather than an inference-time mechanism: at deployment it is removed, and the model routes its own attention to the regions designated by its \texttt{<LOOK>} blocks while running the unmodified forward pass of its backbone.
% We find that this internalized policy is not only cheaper than keeping the bias active at test time, but also more accurate.

To instill this top-down control, we adopt a two-stage training paradigm.
First, we warm-start the model via supervised fine-tuning on a curated dataset demonstrating iterative human-like scanning.
We then refine this policy via Group Relative Policy Optimization (GRPO)~\citep{shao2024deepseekmath}, using a reward system that incentivizes geometrically valid grounding while explicitly penalizing redundant or excessive gazing.
In summary, our key contributions are:

\begin{itemize}
	\item A curated gaze-reasoning dataset: we build an instructional dataset of verified gaze traces designed to train VLMs to transition from passive observation to active, multi-region visual navigation.
	% \item An attention-steering architecture for active vision: we introduce an internal gazing mechanism that dynamically biases the VLM's causal attention mask. This empowers the model to autonomously shift focus from a global view to local details and back, ensuring task-relevant fixation without the computational overhead of adding and processing new visual tokens.
    \item A training signal for active vision: we introduce a gaze bias that modulates the VLM's attention logits over already-encoded visual features during training, teaching the model to autonomously shift focus from a global view to local details and back. It requires no change to the architecture and adds no new visual tokens.
    % NOTE: do we want to highlight this? %MR: I think so, I'll rephrase a bit and put it back
    \item Internalization of the attention bias teaching signal: we show that at deployment time, the attention of the trained model running the unmodified forward pass (with the bias switched off) successfully focuses on the region in its \texttt{<LOOK>} block.
	\item Strong high-resolution reasoning without tool calling and re-encoded visual tokens: applied on 4B backbones, we show that GazeVLM lifts the performance of its base models by about 4 points on HRBench-4k and HRBench-8k, outperforming 7B zoom-in agents by 5 to 12 points, while generating about $5\times$ fewer output tokens.
\end{itemize}

\section{Related Work}

\paragraph{Vision-Language Model reasoning.} 

Considerable research has focused on enhancing the reasoning capabilities of open-weight Vision-Language Models (VLMs) such as Qwen-VL \citep{bai2023qwenvl, wang2024qwen2vl, bai2025qwen25vl, bai2025qwen3vl}, LLaVA \citep{an2025llavaonevision15, liu2024llava15}, and InternVL \citep{zhu2025internvl3, chen2024how}. 
Following the success of Large Language Models (LLMs), a primary strategy has been the adaptation of Chain-of-Thought (CoT) prompting to the multimodal domain, encouraging models to articulate intermediate logical steps \citep{zhang2023multimodalcot}.

To ensure these textual steps remain tethered to the image, a distinct line of work emphasizes grounded reasoning, requiring models to output bounding box coordinates \citep{peng2023kosmos2, you2024ferret, man2025argus} or utilize superimposed visual markers \citep{yang2023setofmark}.
Another prominent direction involves augmenting VLMs with external tools, most notably ``zoom-in'' tools or visual search agents (e.g., V* \citep{wu2024vstar}, Pixel Reasoner \citep{Wangetal2025}, Ground-R1 \citep{Caoetal2025}, DeepEyes \citep{Zhengetal2025}, Visual Perception Token \citep{Yu2025b}).
While effective at capturing fine-grained details, these tool-use frameworks repeatedly crop and pass new image patches through the vision encoder, generating additional visual tokens at each reasoning step.
Closer to our work, \citep{Yu2025b} add Visual Perception Tokens to the vocabulary, which the model generates to trigger its own perception. However, their region selection tokens, still crop and re-encode the selected region, and their vision re-Encoding tokens invoke an additional encoder (DINOv2 or SAM) through a learned projector; both appending new visual tokens.
% To internalize these capabilities into current VLMs, the community has increasingly adopted Reinforcement Learning (RL), with Group Relative Policy Optimization (GRPO) \citep{shao2024deepseekmath} for MLLMs, typically within a multi-stage pipeline of supervised fine-tuning (SFT) followed by RL-driven reasoning refinement.

\paragraph{Attention steering and hallucination in VLMs.} Improving the reasoning of VLMs is closely linked to the reduction of multimodal hallucinations. VLMs are notoriously prone to ``object hallucination'' \citep{rohrbach2018object, biten2022let, li2023evaluating}, where there is an over-reliance on the language prior during generation with respect to the actual visual evidence. Techniques such as Visual Contrastive Decoding (VCD) \citep{leng2024mitigating} attempt to penalize language-driven hallucinations by contrasting output logits between original and explicitly distorted visual inputs. More recently, Attention-Steerable Contrastive Decoding (ASCD) \citep{wang2026ascd} showed that similar hallucination reduction can be achieved more efficiently by directly manipulating the model's internal attention scores, dynamically amplifying logical text-heads while dampening spurious visual tokens. However, these methods are primarily designed to calibrate a single, direct answer and do not account for the dynamic, multi-step nature of complex reasoning. 

In contrast to these test-time interventions, or to the injection of zoom-in visual tokens, our work introduces a training strategy that adds a bias to the decoder's attention logits over already-encoded visual features.
The bias is applied only during training: at deployment the model runs the unmodified forward pass of its backbone.
Inspired by human gaze scanpaths \citep{yarbus1967eye, findlay2003active}, our method enables the VLM to autonomously steer its own visual focus, seamlessly transitioning between global context and local details throughout its reasoning trace. This allows the model to fixate precisely on task-relevant regions without the overhead of physically manipulating or re-encoding the image.

\section{Gaze Reasoning}

\subsection{Gaze actions}

We introduce GazeVLM, a visual reasoning method that enables a VLM to autonomously focus on specific regions of an image, extract relevant information and return to a global view, all through attention-steering of the pre-encoded visual tokens.

\begin{figure}[t]
	% \vskip 0.2in
	% width=\linewidth
	\begin{center}
		\includegraphics[width=\linewidth]{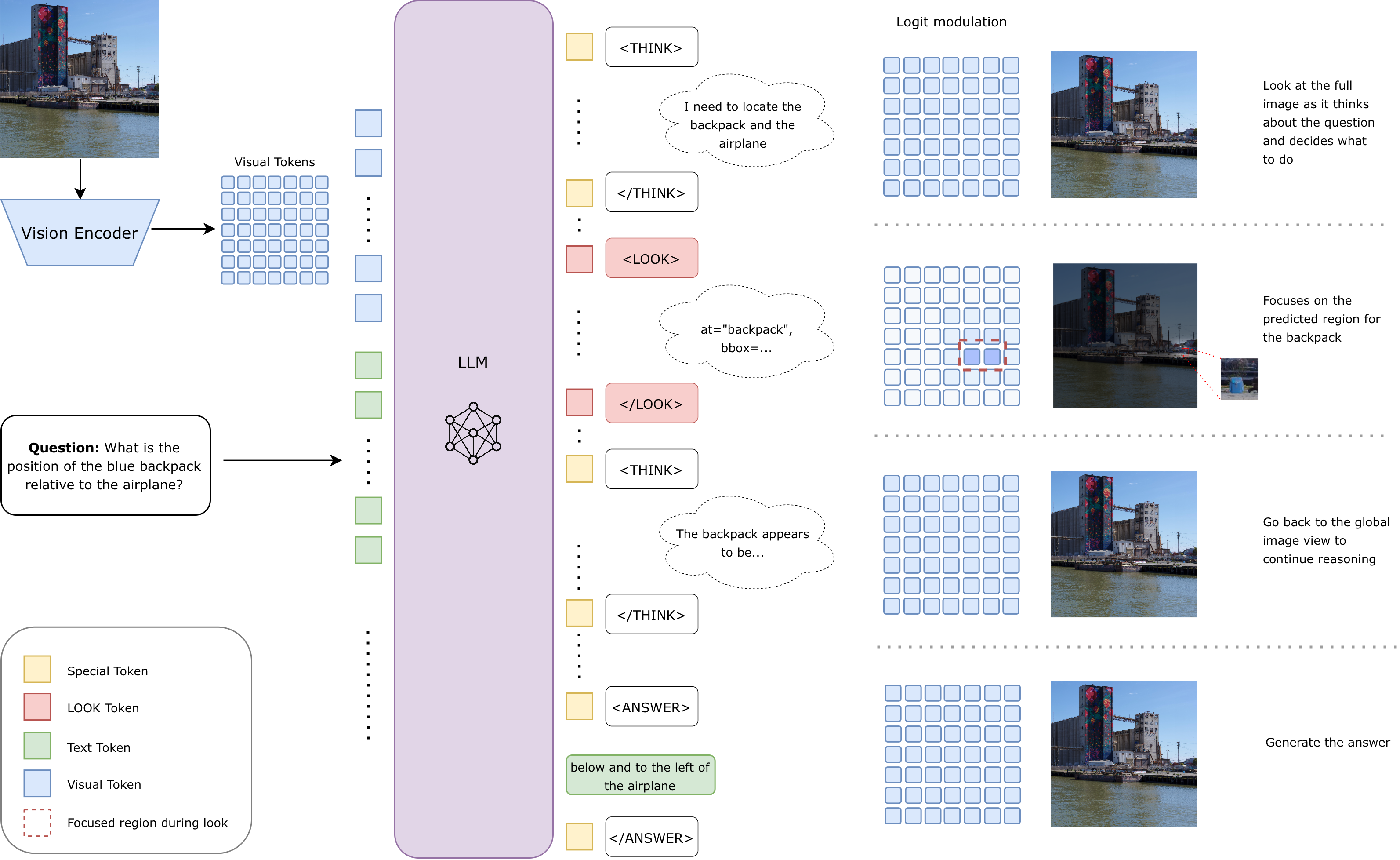}
		\caption{Architectural overview of the GazeVLM reasoning pipeline. Upon receiving a visual query, the model iteratively interleaves logical reasoning \texttt{<THINK>} with active visual navigation. At training, generating a \texttt{<LOOK>} token triggers top-down logit modulation, applying a continuous suppression bias to task-irrelevant visual tokens while preserving the focal region. Once local extraction is complete, the bias is lifted, restoring the global visual context to inform the next reasoning step until reaching the final \texttt{<ANSWER>} generation. The bias is not triggered at deployment.}
		\label{fig:main_arch}
	\end{center}
	\vskip -0.2in
\end{figure}

The architecture of GazeVLM keeps the standard VLM design (e.g. Qwen-VL) unchanged, and comprises a vision encoder, a projector and an LLM.
GazeVLM reasons by interleaving text with \emph{gaze actions} as shown in \Cref{fig:main_arch}, and is trained with an attention bias that ties each action to the model's attention.
Crucially, the bias is a training signal only and is removed at deployment.
Gaze actions are \texttt{<LOOK>} tokens that the model can emit and are annotated with a bounding box indicating the region to focus on.
Concretely, the model emits: \texttt{<LOOK at="[target]" bbox=[x1, y1, x2, y2]> [observation] </LOOK>}.

As it reasons, we allow the model to also generate chains-of-thought with \texttt{<THINK> [thought] </THINK>} at the beginning of the reasoning and between different \texttt{<LOOK>}s.
The model then provides the final answer in \texttt{<ANSWER>}~\textit{final answer}~\texttt{</ANSWER>} tags.

\subsection{Gaze bias as a training signal}

\paragraph{Training and deployment policies.}
Given a vision-language query $(I, x)$, where $I$ is the input image and $x$ is the question, at deployment GazeVLM samples the policy $y_t \sim \pi_\theta(\cdot | I, x, h_t)$ parametrized by model weights $\theta$ with the unmodified forward pass of its base VLM, and where $h_t$ is the generated history of previous steps.

During training, the forward pass is modified by a bias $\beta(B_t)$ (described below) operating on the attention logits, where $B_t$ denotes the bounding boxes of regions of interest according to the trace $h_t$.
Accordingly, the token at the $t$-th generation step is sampled as $y_t \sim \pi^{\beta(B_t)}_\theta(\cdot | I, x, h_t)$ from the biased policy $\pi^{\beta(B_t)}_\theta$.
As a result, the intuition is that during the reasoning process, the policy can decide to focus on key specific areas on the image while generating text grounded on these areas. 
When no focus is active, the model has full access to the global visual context.
The iterative reasoning process terminates when a designated end token is generated.

%Our pipeline would then apply a bias to the LLM's attention over the visual tokens such that regions outside $b$ are dampened, while the targeted region remains fully accessible.
%Our pipeline would then bias the VLM in the decoding steps so that only relevant visual features are visible to encourage the model to ground itself properly reducing irrelevant visual information and noise. 

\paragraph{Attention suppression bias.}

The vision encoder maps input image $I$ to visual tokens $V$. Each visual token $v_i \in V$ is assigned a normalized 2D coordinate $p_i = (c_i, r_i) \in [0, 1]^2$ of the image patch it encodes.

\begin{figure}[t]
	\begin{center}
		\includegraphics[width=0.85\linewidth]{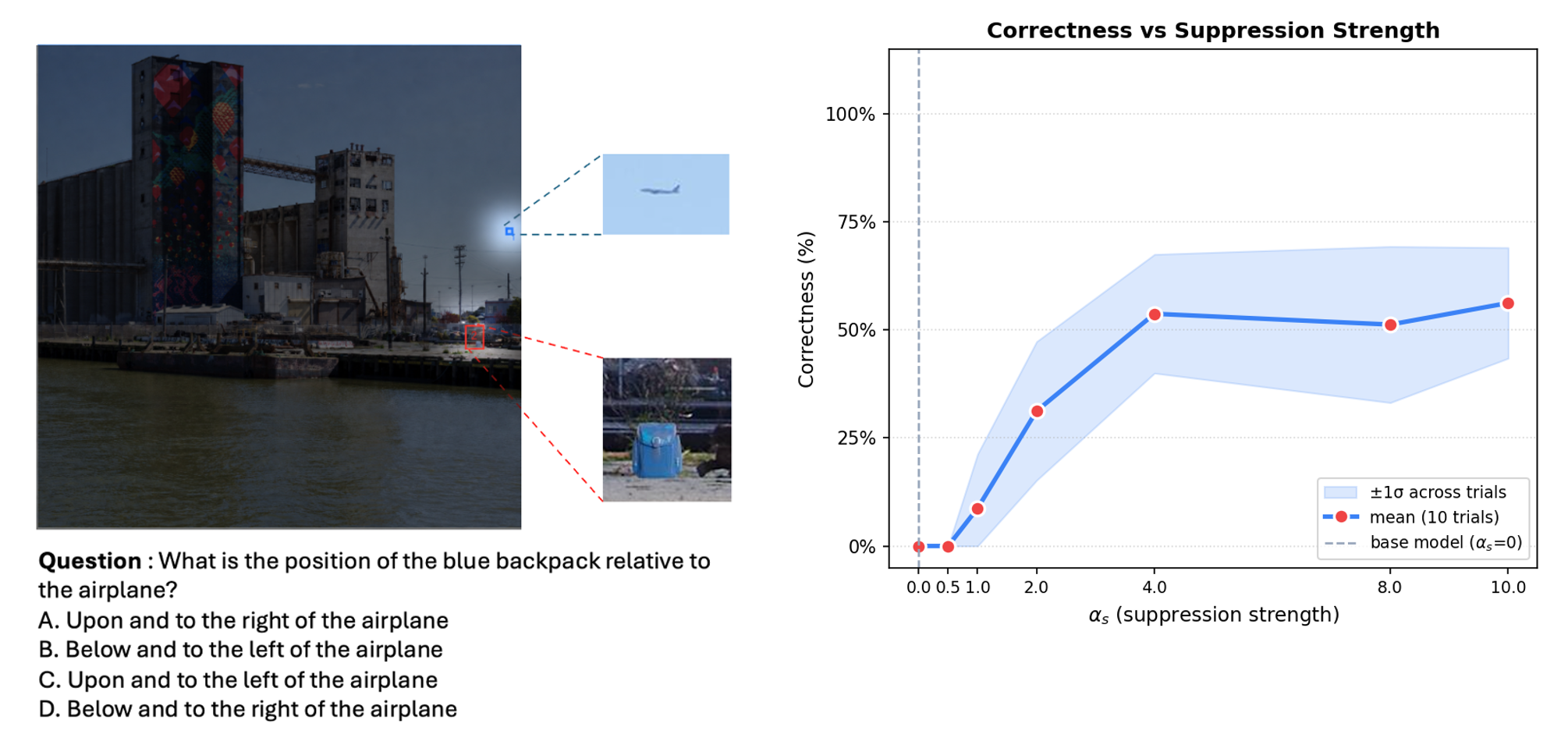}
		\caption{Effect of the suppression bias on one HRBench-4k question. The base Qwen3-VL-4B, without gaze training, is given oracle boxes on the airplane and the backpack; accuracy over 10 trials of 8 rollouts each, as a function of the suppression strength $\alpha_s$ (\Cref{eq:sup_bias_single}).}
		\label{fig:ablate_suppression}
	\end{center}
	\vskip -0.2in
\end{figure}

To enforce focus on a bounding box $b = [x_1, y_1, x_2, y_2]$, instead of re-encoding a cropped image of the box content as other works do, we define $D(p_i, b)$ to be the Euclidean distance from $p_i$ to the box:
\begin{equation}
	D(p_i, b) = \sqrt{\max(x_1 - c_i, c_i - x_2, 0)^2 + \max(y_1 - r_i, r_i - y_2, 0)^2}.
\end{equation}
The suppression bias on token $v_i$ is then
\begin{equation} \label{eq:sup_bias_single}
	\beta_i(b) = -\alpha_s\,\frac{D^2(p_i, b)}{2\,\sigma^2}
\end{equation}
where $\sigma$ sets the spatial falloff and $\alpha_s$ scales the suppression strength.
This bias is added to the attention logits (\Cref{eq:attn_bias}), leaving tokens in the box untouched, while attention to tokens outside is dampened by a Gaussian factor which falls exponentially to zero moving away from the bounding box.
\Cref{fig:ablate_suppression} illustrates the effect of the bias on a single high-resolution example.
Given oracle boxes on the two objects in the question, the base Qwen3-VL-4B, with no gaze training, never answers correctly with unmodified attention ($\alpha_s = 0$); as $\alpha_s$ grows its accuracy rises and levels off at about 55\% from $\alpha_s = 4$, the value we use throughout.

\paragraph{Gaze accumulation and release.}

Questions that compare or relate multiple objects require evidence from multiple regions.
Previous works using zoom-in as a tool retain that evidence by appending to the context the tokens that encode each relevant cropped region.
GazeVLM on the other hand accumulates evidence across multiple gaze actions by aggregating their corresponding bounding boxes.
In particular, after the model has generated a sequence of $K$ target bounding boxes $B = \{b_1, b_2, \ldots, b_K\}$ within its reasoning trace, we apply a bias that is computed as the element-wise maximum across all accumulated gaze actions: $\beta_i(B) := \max_{b \in B} \beta_i(b)$.
This ensures that all visual tokens within any previously inspected area remain fully activated.
As soon as a closing token \texttt{</LOOK>} is generated, the bias is released, so that the model attends to the full image until the next \texttt{<LOOK>} token.

\paragraph{Gaze bias through attention logit modulation.}

The gaze bias is added to the attention logits of all layers and heads of the language decoder.
Given a query $q_t$ and a visual key $k_i$, the pre-softmax attention score $s(q_t, k_i)$ is modulated additively:
\begin{equation} \label{eq:attn_bias}
	\tilde{s}(q_t, k_i) = s(q_t, k_i) + \mathbf{1}_{\textnormal{text}}(q_t) \cdot \beta_i(B_t),
\end{equation}
where $s(q_t, k_i) = \frac{q_t \cdot k_i}{\sqrt{d}}$ with $d$ the head dimension, and $\mathbf{1}_{\textnormal{text}}(q_t)$ is 1 for queries from text tokens and 0 for queries from visual tokens.
This design ensures that the gaze bias only affects how the language model attends to the image.
By isolating the bias to text-to-vision queries, we direct the LLM's focus towards the region inside $B_t$, while preserving the internal representations of the global scene.

Importantly, at deployment we set $\beta \equiv 0$, meaning that \Cref{eq:attn_bias} reduces to the standard attention of the base VLM.

\subsection{Training} \label{sec:training}

We employ a two-stage training pipeline that includes a warm-start SFT training followed by reinforcement learning of the SFT model using GRPO \citep{shao2024deepseekmath}.

\paragraph{Warm-start Instruction Tuning.}
We want to integrate into existing VLMs a new reasoning paradigm where the model can reason, focus on an area through our gazing mechanism, and then zoom-out into the global image to continue to reason until it reaches the final answer. To equip an existing Qwen model such as Qwen3-VL-Instruct \citep{bai2025qwen3vl} to follow this pattern, which it learns during RL, we first need to help the model understand how it should do that through SFT (Supervised Fine-Tuning).

\paragraph{Dataset generation.} \label{par:dataset}

We design our data generation pipeline so that our data (reasoning traces) display the gazing mechanism that we want to teach the model to follow. Our specialized dataset is sourced from: GQA \citep{hudson2019gqa}, ChartQA \citep{masry2022chartqa}, PlotQA \citep{methani2020plotqa}, and InfoVQA \citep{mathew2022infographicvqa}. We leverage Qwen3-VL-235B-A22B as our expert model to synthesize trajectories that explicitly utilize our gazing mechanism. Comprehensive details on dataset generation and heuristic filtering are provided in \Cref{app:dataset}.

We train the base model on these synthesized traces using standard cross-entropy loss, activating the continuous suppression bias during the forward pass so the model learns the visual consequences of its \texttt{<LOOK>} coordinates.

\paragraph{Learning gaze actions with Reinforcement Learning.}

Starting from the SFT model, we optimize the policy with GRPO \citep{shao2024deepseekmath}.
Inspired by other works in RL \citep{Wangetal2025, pathak2017curiosity}, we craft a gaze-driven reward to encourage the model to \textit{look} at relevant areas of the image during its reasoning.
More precisely, for a given sample consisting of an image $I$, a question $x$, and a ground truth answer $y^*$, we sample a group of $G$ distinct completions $\{c_1, c_2, \ldots, c_G\}$ under the policy $\pi^\beta_\theta$ with the bias active.
Each completion $c_i$ is then scored by a composite reward function $R(c_i)$ designed to enforce answer correctness, visual navigation, and grounding.
For a completion $c$, we specify the total reward as
\begin{equation} \label{eq:reward}
	R(c) = R_{\textnormal{correct}}(c) + R_{\textnormal{format}}(c) + R_{\textnormal{bbox}}(c) + R_{\textnormal{overlap}}(c) + R_{\textnormal{excess}}(c) + R_{\textnormal{len}}(c).
\end{equation}

A comprehensive breakdown of each specific reward component, the penalty thresholds, and the formal GRPO objective are detailed in \Cref{app:grpo_details}.

\section{Experiments}

\paragraph{Experimental Setup and Baselines.} We build GazeVLM upon the Qwen3-VL-4B-Instruct architecture. The model is trained on our highly curated, gaze reasoning traces and optimized via GRPO. We evaluate GazeVLM against multiple VQA benchmarks spanning complex chart and logical reasoning (MathVista \citep{lu2024mathvista}, ChartQA \citep{masry2022chartqa}), general reasoning (MMBench \citep{liu2024mmbench}, MMStar \citep{chen2024mmstar}, CV-Bench \citep{tong2024cambrian1}), and high-resolution benchmarks (HRBench-4k, HRBench-8k \citep{wang2025divide}). We compare our results against a wide range of models, including closed-source models, vanilla open-source VLMs, SFT-trained reasoning models (Vision-R1-LlamaV-CI \citep{huang2025visionr1}, SPARC \citep{avogaro2026sparc}), and state-of-the-art RL-trained agents (Vision-R1 \citep{huang2025visionr1}, Pixel-Reasoner \citep{Wangetal2025}, DeepEyes \citep{Zhengetal2025}, and Ground-R1 \citep{Caoetal2025}). Full details regarding dataset sizes, training settings, and baseline evaluation protocols are provided in \Cref{app:dataset} and \Cref{app:exp_details}.

\paragraph{Main results.}

\begin{figure}[t]
	\centering
	
	\begin{subfigure}{0.4\linewidth}\centering
		\includegraphics[width=\linewidth, trim=0 0 0 600px, clip]{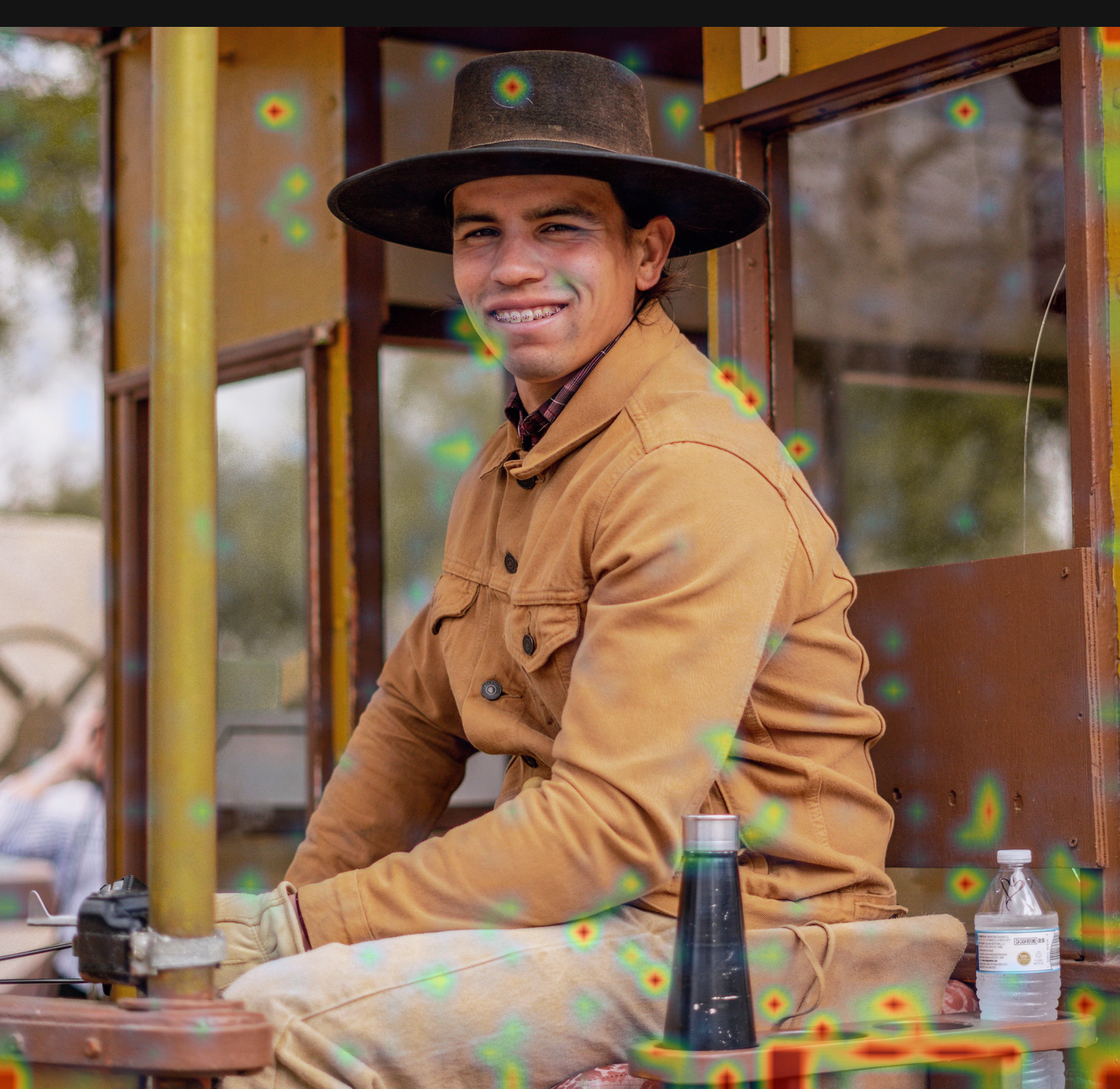}
		\caption{Vanilla Qwen3-VL-4B-Instruct}
	\end{subfigure}
	\begin{subfigure}{0.4\linewidth}\centering
		\includegraphics[width=\linewidth, trim=0 0 0 600px, clip]{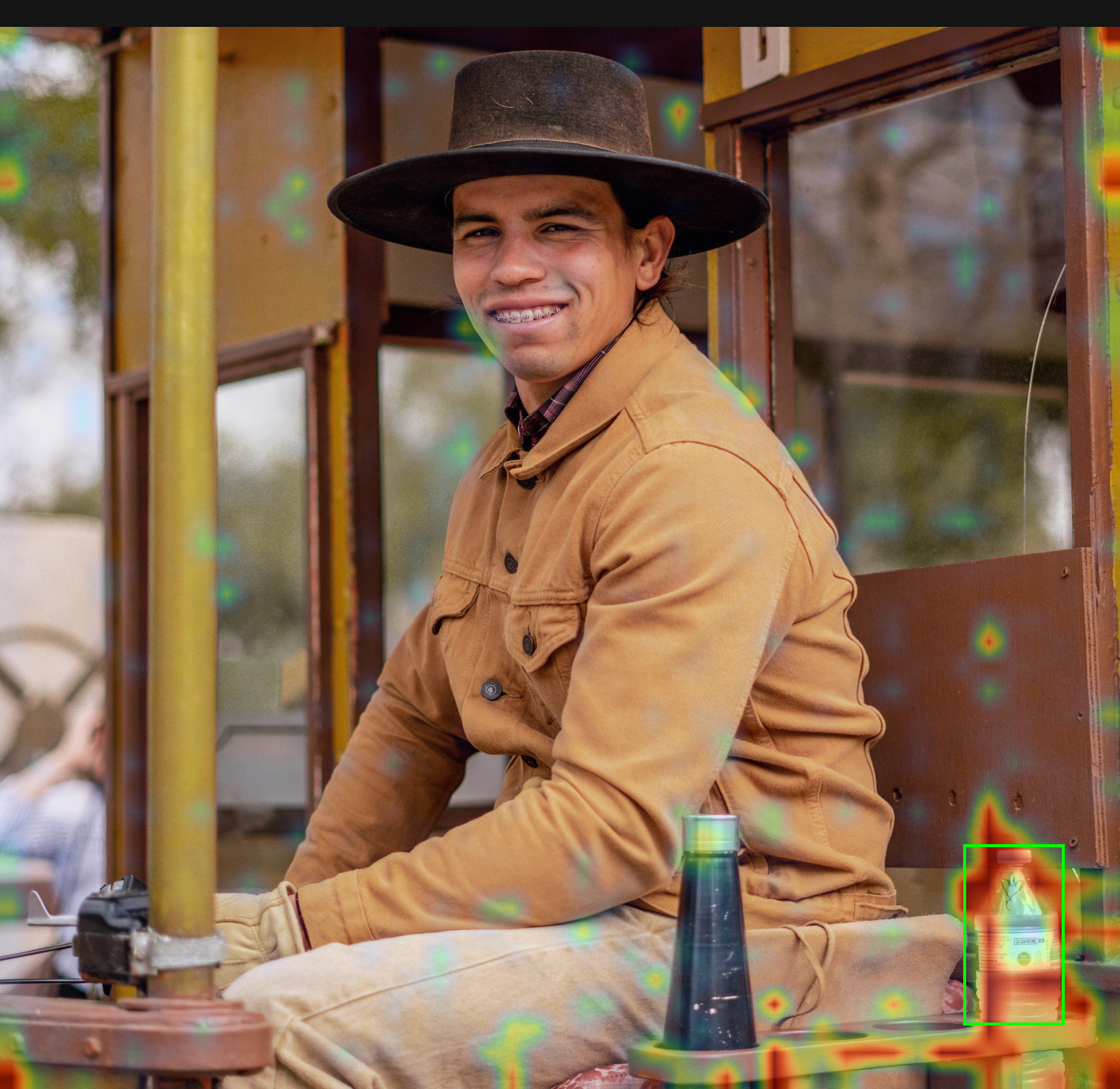}
		\caption{GazeVLM}
	\end{subfigure}
	\caption{%
		\textbf{Internalized visual focus of GazeVLM.}
		We compare \emph{(a)} vanilla Qwen3-VL-4B-Instruct and \emph{(b)} GazeVLM on sample 32 of HRBench-4k, on the question ``\emph{Where is the water bottle placed relative to the person in the image?}''.
		Both models run the same unmodified forward pass, as GazeVLM at \emph{(b)} reflects the attention pattern internalized through training. The green box is the region GazeVLM designated in its own \texttt{<LOOK>} block.
		Heatmaps show the attention from each generated trace token to the visual tokens, averaged over tokens, decoder layers, and heads. Training shifts attention from a diffuse, content-driven distribution to one tightly focused on the region the model identifies as task-relevant.}
	\label{fig:learned-attention}
\end{figure}

\Cref{fig:learned-attention} provides a qualitative illustration of how GazeVLM training reshapes visual attention, on a representative sample from HRBench-4k.
We visualize the decoder attention of the Qwen3-VL-4B-Instruct base model (left) and of its GazeVLM-finetuned counterpart (right), averaged over the tokens generated in response to the query \emph{``Where is the water bottle placed relative to the person in the image?''}.
The base model's attention is diffuse, and it answers incorrectly that \emph{``The bottle is in front of the person''}.
GazeVLM, by contrast, focuses tightly on the water bottle, within the bounding box that it emitted in its own \texttt{<LOOK>} trace, and answers correctly that \emph{``The bottle is placed to the right of the person''}.

Beyond this qualitative illustration, \Cref{tab:main_results} reports our main results aggregated across a suite of VQA benchmarks.
To rigorously isolate the contribution of our gaze mechanism from that of our data curation, we highlight two reference baselines.
The first is the off-the-shelf base model Qwen3-VL-4B.
The second is a computationally matched passive grounding baseline, denoted GazeVLM (w/o gaze bias): trained on the same curated dataset and the same GRPO pipeline, but operating entirely passively, emitting text and bounding box coordinates without modulating the underlying attention logits, in the spirit of grounded reasoning works such as Argus \citep{man2025argus}.

This passive grounding baseline is already competitive, outperforming RL-trained agentic VLMs such as DeepEyes \citep{Zhengetal2025} on HRBench-8k (74.4 vs.\ 72.6) and validating the quality of our curated traces.
Activating our core architectural contribution, the dynamic top-down attention suppression, then delivers a further performance leap on top of this strong baseline, with consistent gains across chart and math reasoning, general reasoning, and high-resolution benchmarks.
On complex logical reasoning, GazeVLM achieves state-of-the-art results within its parameter class, pushing MathVista to 73.3 ($+5.2\%$ over the passive baseline) and MMBench to 89.2.
The gains are most pronounced on the high-resolution HRBench datasets, where GazeVLM scores 83.4 on HRBench-4k ($+3.9$ over the base model) and 78.0 on HRBench-8k ($+4.4$).

\begin{table*}[t]
	\centering
	\caption{
		Comprehensive evaluation of GazeVLM against state-of-the-art vision-language models. We group benchmarks by their primary cognitive demands.
        % The $\Delta$ row illustrates the absolute performance leap achieved by our GazeVLM
        % compared to the standard Qwen3-VL-4B base model.
		% \colorbox{bestcolor}{\textbf{Best}} and 
		% \colorbox{secondcolor}{\underline{second best}} results are highlighted.
	}
	\label{tab:main_results}
	\vspace{4pt}
	\resizebox{\textwidth}{!}{%
		\begin{tabular}{l l cc ccc cc}
			\toprule
			& & \multicolumn{2}{c}{\textit{Chart \& Math Reasoning}} 
			& \multicolumn{3}{c}{\textit{General Reasoning \& Vision critical}} 
			& \multicolumn{2}{c}{\textit{High-Resolution}} \\
			% & \multicolumn{1}{c}{\textit{Perception}} \\
			\cmidrule(lr){3-4} \cmidrule(lr){5-7} \cmidrule(lr){8-9} 
			% \cmidrule(lr){10-10}
			& \textbf{Method}
			& \rotatebox{45}{\textbf{ChartQA}} 
			& \rotatebox{45}{\textbf{MathVista}} 
			& \rotatebox{45}{\textbf{MMBench}}
			& \rotatebox{45}{\textbf{MMStar}}
			& \rotatebox{45}{\textbf{CV-Bench}} 
			% & \rotatebox{45}{\textbf{V*}} 
			& \rotatebox{45}{\textbf{HRBench-4k}} 
			& \rotatebox{45}{\textbf{HRBench-8k}} \\
			% & \rotatebox{45}{\textbf{InfoVQA}} \\
			\midrule
			
			\rowcolor{grouphead}
			& \multicolumn{8}{l}{\small\textit{Closed-source models}} \\
			& OpenAI-o1     & - & 73.9 & - & - & - & - & - \\
			& GPT-4o     & - & - & - & - & 66.0 &  59.0 & 55.5 \\
			& Gemini2.5-Flash-Lite & - & 70.3 & - & - & 64.9 & 67.2 & - \\
			
			% --- Group: Vanilla VLMs ---
			\rowcolor{grouphead}
			& \multicolumn{8}{l}{\small\textit{Vanilla Open-source VLMs}} \\
			& Qwen2.5-VL-7B     & 83.3 & 67.9 & 83.8 & 57.9 & 73.7 & 68.5 & 59.5 \\
			% & InternVL2.5-8B     & - & - & - & - & - & - & - & - \\
			& Qwen3-VL-4B   & 83.2 & 71.7 & 87.7 & 67.3 & 83.4 & 79.5 & 73.6  \\
			& Qwen3.5-4B-Inst   & 78.8 & 55.7 & 83.8 & 58.3 & 81.1 & 78.6 & 74.0 \\
			& Qwen3.5-4B-Think   & 83.1 & 68.8 & 88.0 & 65.4 & 84.5 & 83.0 & 78.4 \\
			\midrule
			
			% --- Group: Reasoning VLMs (SFT) ---
			\rowcolor{grouphead}
			& \multicolumn{8}{l}{\small\textit{Reasoning VLMs (SFT-based)}} \\
			% & LLaVA-o1-7B        & - & - & - & - & - & - & - & - \\
			% & Qwen3-VL-8B-Think  & - & - & - & - & - & - & - & - \\
			
			& Vision-R1-LlamaV-CI-11B & 83.9 & 62.7 & - & 61.4 & - & - & - \\
			& SPARC-4B \citep{avogaro2026sparc} & - & - & - & - & - & 70.5 & 68.4 \\
			% \citep{huang2025visionr1} 
			\midrule
			
			% --- Group: RL-trained VLMs ---
			\rowcolor{grouphead}
			& \multicolumn{8}{l}{\small\textit{RL-trained VLMs}} \\

			& Vision-R1-7B \citep{huang2025visionr1}  & - & 73.5 & - & - & - & - & - \\
			& Pixel-Reasoner-7B \citep{Wangetal2025}  & - & - & - & - & - & 74.0 & 66.9 \\
			& DeepEyes  \citep{Zhengetal2025}       & - & 70.1 & - & - & - & 75.1 & 72.6 \\
			& Ground-R1-7B \citep{Caoetal2025}      & - & - & - & - & - & 75.0 & 71.1 \\
			\midrule
			
			% --- Our method ---
			% & \textbf{SFT (w/o gaze)} & \textbf{51.3} & \textbf{75.2} & \textbf{86.3} & \textbf{67.4} & \textbf{00.0} & \textbf{00.0} & \textbf{00.0} & \textbf{83.4} \\

			% \rowcolor{ourscolor}
			& GazeVLM (w/o gaze bias) & 85.5 & 68.1 & 88.7 & 67.1 & 84.5 & 79.0 & 74.4  \\
			
			% \midrule
			\rowcolor{ourscolor}
			& \textbf{GazeVLM (Ours)} & \textbf{85.8} & \textbf{73.3} & \textbf{89.2} & \textbf{69.0} & \textbf{85.4} & \textbf{83.4} & \textbf{78.0}  \\ % 78.7
			
			\rowcolor{ourscolor}
			& $\Delta$ (vs Qwen3-VL-4B) & $+2.6$ & $+1.6$ & $+1.5$ & $+1.7$ & $+2.0$ & $+3.9$ & $+4.4$ \\

            \midrule
			
			\rowcolor{ourscolor}
			& \textbf{GazeVLM-Qwen3.5 (Ours)} & \textbf{85.8} & \textbf{77.4} & \textbf{90.0} & \textbf{71.2} & \textbf{86.2} & \textbf{86.9} & \textbf{82.1}  \\ % 78.7
            \rowcolor{ourscolor}
			& $\Delta$ (vs Qwen3.5-4B-Think) & $+2.7$ & $+8.6$ & $+2.0$ & $+5.8$ & $+1.7$ & $+3.9$ & $+3.7$ \\
			\bottomrule
		\end{tabular}
	}%
	\vspace{-8pt}
\end{table*}

\section{Mechanism analysis}

This section tests whether the gains come from the gaze bias, and whether the trained model's attention follows its own gaze actions once the bias is removed.

\paragraph{Gaze bias at training.}
The passive grounding baseline, GazeVLM (w/o gaze bias), shares every ingredient of GazeVLM except the bias: the Qwen3-VL-235B-A22B teacher traces, the prompts, the box supervision, the SFT and GRPO recipe and the reward.
Like standard grounding models \citep{peng2023kosmos2, you2024ferret, man2025argus}, it therefore emits boxes that do not affect its attention.
After GRPO it trails GazeVLM on every benchmark in \Cref{tab:ablation_gaze_importance}, by 5.2 points on MathVista and by 4.4 and 3.6 on HRBench-4k and HRBench-8k, where it stays within one point of the base model.
GazeVLM and the baseline start GRPO from their own SFT checkpoints, trained with and without the bias respectively, but the gap is not inherited from SFT: after SFT alone, training with the bias brings no gain (68.4 vs.\ 68.7 on MathVista, 80.0 vs.\ 79.8 on HRBench-4k, 73.5 vs.\ 76.0 on HRBench-8k), and the gap opens only during GRPO.

\paragraph{Gaze bias at inference.}
Applying the bias at inference as well brings no gain: with it, GazeVLM scores 69.4 on MathVista, 82.5 on HRBench-4k and 76.5 on HRBench-8k, against 73.3, 83.4 and 78.0 without it, which is expected if the trained model already attends where the bias would steer it.
All GazeVLM results are reported with the bias off, i.e.\ with unmodified forward pass of the VLM.
% Below, we test directly whether the model's attention follows its own gaze actions in this setting.

\begin{table*}[t]
	\centering
	\caption{
        Effect of training with gaze bias after SFT and GRPO. Models are evaluated with bias off.
	}
	\label{tab:ablation_gaze_importance}
	\vspace{4pt}
	\resizebox{0.95\textwidth}{!}{%
		\begin{tabular}{l c ccccc}
			\toprule
			& &
			\multicolumn{5}{c}{\textit{Benchmarks}} \\
			\cmidrule(lr){3-7}
			\textbf{Training} &
			\rotatebox{0}{\small\textbf{Train w/ Gaze}} &
			% \small\textbf{MMMU} &
			\small\textbf{MathVista} &
			\small\textbf{MMStar} &
			% \small\textbf{GQA} &
			\small\textbf{ChartQA} &
			% \small\textbf{CharXiv} &
			% \small\textbf{VStar} &
			\small\textbf{HRBench-4k} &
			\small\textbf{HRBench-8k} \\
			\midrule

			Qwen3-VL-4B & \no & 71.7 & 67.3 & 83.2 & 79.5 & 73.6 \\
			% qwen3.5-ins. &\no & \no  & \no  & 56.3 & 60.1 & 79.2 & - & - \\
			
			% --- SFT block ---
			\midrule
			\multirow{2}{*}{\textit{SFT}}
			& \no  & 68.7 & 65.9 & 83.7 & 79.8 & 76.0 \\[2pt]
			& \yes & 68.4 & 65.7 & 83.8 & 80.0 & 73.5 \\

			\midrule
			
			% --- GRPO block ---
			\multirow{2}{*}{\textit{GRPO}}
			& \no  & 68.1 & 67.1 & 85.5 & 79.0 & 74.4\\[2pt]

            % \rowcolor{ourscolor}
			& \yes & \textbf{73.3} & \textbf{69.0} & \textbf{85.8} & \textbf{83.4} & \textbf{78.0}\\
			
			\bottomrule
		\end{tabular}
	}%
	
	% \vspace{6pt}
	
	% \vspace{-6pt}
	
\end{table*}

\begin{table}[t]
\small
\centering
\caption{Attention--region agreement on 200 HRBench samples.
IoU between the top-$k$ attended visual tokens and the $k$ tokens inside the emitted \texttt{<LOOK>} box (mean $\pm$ standard error).
The base model is evaluated on GazeVLM traces.
The condition ``GazeVLM + bias'' artificially applies the bias to check its limited additional contribution.}
\label{tab:routing}
\begin{tabular}{lc}
\toprule
\textbf{Model} & \textbf{IoU} \\
\midrule
Qwen3-VL-4B (base)   & 0.130 $\pm$ 0.009 \\[2pt]
GazeVLM              & 0.360 $\pm$ 0.018 \\[2pt]
GazeVLM \textbf{+ bias}       & 0.440 $\pm$ 0.018 \\
\bottomrule
\end{tabular}
\end{table}

\paragraph{Is the teaching signal internalized?}
Here we show that the teaching signal is \emph{internalized}, i.e.\ after training and with the bias off, the model's own gaze actions shift its attention to the named region as the bias would, and the model relies on them to answer successfully.
We test this in two steps: by showing that attention agrees with the emitted box, and verifying that accuracy degrades close to base performance when the gaze actions are removed.

For each \texttt{<LOOK>} block with box $b$, let $\mathcal{R}$ be the $k$ visual tokens in $b$, and $\mathcal{H}$ the $k$ visual tokens receiving the most attention from the text tokens of the block, averaged over layers and heads.
We report $\mathrm{IoU}(\mathcal{H}, \mathcal{R}) = |\mathcal{H} \cap \mathcal{R}| / |\mathcal{H} \cup \mathcal{R}|$, which is 1 when the most-attended tokens are exactly those in the box.
For the base model, which emits no gaze actions, we feed it GazeVLM's traces and measure against the same boxes.
On 200 HRBench samples (100 each from 4k and 8k), \Cref{tab:routing} shows that the base model reaches an IoU of 0.130, while GazeVLM reaches a much higher IoU of 0.360, indicating that the model's attention concentrates on the named relevant region.
Artificially applying the bias on top (the GazeVLM + bias condition in \Cref{tab:routing}) raises the IoU only to 0.440, meaning that GazeVLM has almost completely internalized the effect of the external bias.

Since the model produces both the box and the attention, we investigate whether the gaze action drives the model's behavior by intervening on them.
Specifically, we run GazeVLM with \texttt{<LOOK>} blocks disallowed at inference, so that it can generate \texttt{<THINK>} and \texttt{<ANSWER>} tags to reason, but is unable to produce an explicit observation step. 
Without \texttt{<LOOK>}, the accuracy drops from 82.0 to 76.0 on HRBench-4k and from 76.0 to 71.0 on HRBench-8k (\Cref{tab:look_vs_no_look}).
A careful inspection of the additional mistakes confirms that they mostly involve a failure in extracting fine details.

\begin{table}[t]
\small
\centering
\caption{Effect of removing \texttt{<LOOK>} blocks at inference. Accuracy on HRBench (100 samples each for 4k and 8k) with greedy decoding is reported, showing a drop when \texttt{<LOOK>} blocks are removed.
}
\label{tab:look_vs_no_look}
\begin{tabular}{lcc}
\toprule
\textbf{Model} & \textbf{HRBench-4k} & \textbf{HRBench-8k} \\
\midrule
GazeVLM   & \textbf{82.0} & \textbf{76.0} \\[2pt]
GazeVLM (w/o \texttt{<LOOK>}) & 76.0 & 71.0 \\
\bottomrule
\end{tabular}
\end{table}

\paragraph{Computational efficiency.}
\Cref{tab:efficiency} compares GazeVLM with DeepEyes \citep{Zhengetal2025}, an RL-trained agent that zooms by cropping and re-encoding image regions, on HRBench-4k and HRBench-8k.
Each DeepEyes zoom appends newly encoded visual tokens to the context, and its traces average 1{,}325 output tokens on HRBench-4k; GazeVLM encodes the image once and produces 266 output tokens per trace, about 5$\times$ fewer.
GazeVLM also uses fewer visual actions (1.19 \texttt{<LOOK>}s vs.\ 1.81 zooms per trace) while scoring 8.3 points higher on HRBench-4k and 5.4 on HRBench-8k.

\begin{table}[t]
	\centering
	\caption{Comparison of computational efficiency and visual routing behavior. Token generation and active vision interventions of GazeVLM (internal foveation) are compared against DeepEyes (external cropping) on high-resolution benchmarks. We also provide token counts for the base model. The \textit{efficiency gain} refers to the comparison between GazeVLM and DeepEyes.}
	% \vspace{-3pt}
	\vspace{2pt}
	\label{tab:efficiency}
	\resizebox{\textwidth}{!}{%
		
		\begin{tabular}{lcccccc}
			
			\toprule
			& \multicolumn{3}{c}{\textit{HRBench-4k}} & \multicolumn{3}{c}{\textit{HRBench-8k}} \\
			
			\cmidrule(lr){2-4} \cmidrule(lr){5-7}
			
			\textbf{Method} & \textbf{Tokens / Trace} $\downarrow$ & \textbf{Calls / Trace} $\downarrow$ & \textbf{Accuracy} $\uparrow$ & \textbf{Tokens / Trace} $\downarrow$ & \textbf{Calls / Trace} $\downarrow$ & \textbf{Accuracy} $\uparrow$ \\
            \midrule
			
            Qwen3-VL-4B-Instruct & 113.5 & 0.0 & 79.5 & 125.2 & 0.0 & 73.6 \\
            
            \midrule
			
			DeepEyes (\texttt{ZOOM}) & 1325 & 1.81 & 75.1 & 1268 & 1.71 & 72.6 \\[4pt]
			
			\textbf{GazeVLM (\texttt{<LOOK>})} & \textbf{265.9} & \textbf{1.19} & \textbf{83.4} & \textbf{292.9} & \textbf{1.19} & \textbf{78.0} \\
			\midrule
			
			\textit{Efficiency Gain} & \textit{4.98$\times$} & \textit{1.52$\times$} & \textit{+8.3\%} & \textit{4.33$\times$} & \textit{1.43$\times$} & \textit{+5.4\%} \\
			
			\bottomrule
		\end{tabular}%
	}
	
	% \vspace{2pt}
	
\end{table}

\section{Discussion and Conclusion}

GazeVLM departs from prevailing visual reasoning paradigms by learning top-down control over its own attention logits rather than invoking external cropping tools, implementing active vision \emph{inside} the model as an operation on already-encoded visual features.
A single attention-modulation primitive, gated by \texttt{<LOOK>}/\texttt{</LOOK>} tokens the policy emits itself, drives global-to-local-to-global scanning with no tool calls or extra visual tokens, and is trained end-to-end with GRPO under a geometrically grounded reward.
Notably, once trained the model performs best at inference with the explicit bias turned off, consistent with the attention prior having been internalized into the weights and acted as a training-time regularizer.
Empirically, GazeVLM matches or beats other RL-trained agentic VLMs across general, chart, math, and high-resolution benchmarks, with the largest gains where foveal control matters most (HRBench-4k +3.9, HRBench-8k +4.4), while emitting about $5\times$ fewer output tokens per trace than zoom-in-based agents.
More broadly, GazeVLM shows that a model can be taught self-directed control over its own visual processing, echoing the top-down routing of human gaze to task-relevant regions, and is a step toward broader meta-cognitive control of a model's own computations.

\paragraph{Limitations and broader impact.}
Because GazeVLM modulates attention over pre-encoded features and never re-encodes patches, when targets occupy only a handful of pixels at native resolution (e.g., V*~\citep{wu2024vstar}) the signal is lost during tokenization and no reweighting recovers it; a hybrid policy that falls back to re-encoding when foveation is insufficient is a natural extension.
Our curated data also retains only successful \texttt{<LOOK>} traces, biasing the policy toward gazing even when a global view suffices; mixing in non-gazing traces and rewarding global answers when adequate should improve robustness.
The explicit gaze trace offers built-in interpretability by exposing where the model attended at each step, but sharper localization is dual-use (document understanding and accessibility on one hand, surveillance and biometric tracking on the other), and \texttt{<LOOK>} blocks should be treated as an auditable reasoning aid rather than a certified attestation in high-stakes settings.

\newpage

\bibliography{iclr2027_conference}

@article{yu2025b,
  title={Introducing visual perception token into multimodal large language model},
  author={Yu, Runpeng and Ma, Xinyin and Wang, Xinchao},
  journal={arXiv preprint arXiv:2502.17425},
  year={2025}
}

@book{findlay2003active,
  title={Active vision: The psychology of looking and seeing},
  author={Findlay, John M and Gilchrist, Iain D},
  number={37},
  year={2003},
  publisher={Oxford University Press}
}

@book{yarbus1967eye,
  title={Eye movements and vision},
  author={Yarbus, Alfred L},
  year={1967},
  publisher={Plenum Press}
}

@article{corbetta2002control,
  title={Control of goal-directed and stimulus-driven attention in the brain},
  author={Corbetta, Maurizio and Shulman, Gordon L},
  journal={Nature reviews neuroscience},
  volume={3},
  number={3},
  pages={201--215},
  year={2002},
  publisher={Nature Publishing Group}
}

@article{rensink2000dynamic,
  title={The dynamic representation of scenes},
  author={Rensink, Ronald A},
  journal={Visual cognition},
  volume={7},
  number={1-3},
  pages={17--42},
  year={2000},
  publisher={Taylor \& Francis}
}

@article{shao2024deepseekmath,
  title={DeepSeekMath: Pushing the Limits of Mathematical Reasoning in Open Language Models},
  author={Shao, Zhihong and Wang, Peiyi and Zhu, Qihao and Xu, Runxin and Song, Junxiao and Bi, Xiao and Zhang, Haowei and Zhang, Mingchuan and Li, Y.K. and Wu, Yang and Guo, Daya},
  journal={arXiv preprint arXiv:2402.03300},
  year={2024},
  url={https://arxiv.org/abs/2402.03300}
}

@inproceedings{bae2026online,
	title={Online difficulty filtering for reasoning oriented reinforcement learning},
	author={Bae, Sanghwan and Hong, Jiwoo and Lee, Min Young and Kim, Hanbyul and Nam, JeongYeon and Kwak, Donghyun},
	booktitle={Proceedings of the 19th Conference of the European Chapter of the Association for Computational Linguistics (Volume 1: Long Papers)},
	pages={700--719},
	year={2026}
}

@inproceedings{pathak2017curiosity,
  title={Curiosity-driven exploration by self-supervised prediction},
  author={Pathak, Deepak and Agrawal, Pulkit and Efros, Alexei A and Darrell, Trevor},
  booktitle={International conference on machine learning},
  pages={2778--2787},
  year={2017},
  organization={PMLR}
}

@article{Wangetal2025,
  title   = {Pixel Reasoner: Incentivizing Pixel-Space Reasoning with Curiosity-Driven Reinforcement Learning},
  author={Su, Alex and Wang, Haozhe and Ren, Weiming and Lin, Fangzhen and Chen, Wenhu},
  journal={Advances in Neural Information Processing Systems},
  volume={38},
  pages={8222--8251},
  year={2026}
}

@article{zhang2025adaptive,
  title={Adaptive Chain-of-Focus Reasoning via Dynamic Visual Search and Zooming for Efficient VLMs},
  author={Zhang, Xintong and Gao, Zhi and Zhang, Bofei and Li, Pengxiang and Zhang, Xiaowen and Liu, Yang and Yuan, Tao and Wu, Yuwei and Jia, Yunde and Zhu, Song-Chun and others},
  journal={arXiv preprint arXiv:2505.15436},
  year={2025},
  url={https://arxiv.org/abs/2505.15436}
}

@inproceedings{huang2025visionr1,
  title={Vision-R1: Incentivizing reasoning capability in multimodal large language models},
  author={Huang, Wenxuan and Jia, Bohan and Cao, Shaosheng and Ye, Zheyu and Xu, Zhe and Hu, Yao and Lin, Shaohui and others},
  booktitle={International Conference on Learning Representations},
  volume={2026},
  pages={63794--63812},
  year={2026}
}

@article{Caoetal2025,
  title   = {Ground-R1: Incentivizing Grounded Visual Reasoning via Reinforcement Learning},
  author  = {Cao, Meng and Zhao, Haoze and Zhang, Can and Chang, Xiaojun and Reid, Ian and Liang, Xiaodan},
  journal = {arXiv preprint arXiv:2505.20272},
  year    = {2025},
  doi     = {10.48550/arxiv.2505.20272}
}

@inproceedings{Zhengetal2025,
  title={DeepEyes: Incentivizing "Thinking with Images" via Reinforcement Learning},
  author={Zheng, Ziwei and Yang, Minghao and Hong, Jack and Zhao, Chenxiao and Xu, Guohai and Yang, Le and Shen, Chao},
  booktitle={International Conference on Learning Representations},
  volume={2026},
  pages={126775--126798},
  year={2026}
}

@inproceedings{leng2024mitigating,
    title={Mitigating object hallucinations in large vision-language models through visual contrastive decoding},
    author={Leng, Sicong and Zhang, Hang and Chen, Guanzheng and Li, Xin and Lu, Shijian and Miao, Chunyan and Bing, Lidong},
    booktitle={2024 IEEE/CVF Conference on Computer Vision and Pattern Recognition (CVPR)},
    pages={13872--13882},
    year={2024},
    organization={IEEE}
}

@inproceedings{wang2026ascd,
  title={ASCD: Attention-Steerable Contrastive Decoding for Reducing Hallucination in MLLM},
  author={Wang, Yujun and Bi, Jinhe and Pirk, Soren and Ma, Yunpu and others},
  booktitle={Proceedings of the AAAI Conference on Artificial Intelligence},
  volume={40},
  number={12},
  pages={10306--10314},
  year={2026}
}

@inproceedings{avogaro2026sparc,
    title={{SPARC}: Separating Perception And Reasoning Circuits for Test-time Scaling of {VLM}s},
    author={Niccol{\`o} Avogaro and Nayanika Debnath and Li Mi and Thomas Frick and Junling Wang and Zexue He and Hang Hua and Konrad Schindler and Mattia Rigotti},
    booktitle={Forty-third International Conference on Machine Learning},
    year={2026},
    url={https://openreview.net/forum?id=c0lJ3rOBPu}
}

@inproceedings{biten2022let,
  title={Let there be a clock on the beach: Reducing object hallucination in image captioning},
  author={Biten, Ali Furkan and G{\'o}mez, Llu{\'\i}s and Karatzas, Dimosthenis},
  booktitle={2022 IEEE/CVF Winter Conference on Applications of Computer Vision (WACV)},
  pages={2473--2482},
  year={2022},
  organization={IEEE}
}

@inproceedings{li2023evaluating,
  title={Evaluating Object Hallucination in Large Vision-Language Models},
  author={Li, Yifan and Du, Yifan and Zhou, Kun and Wang, Jinpeng and Zhao, Xin and Wen, Ji-Rong},
  booktitle={Proceedings of the 2023 Conference on Empirical Methods in Natural Language Processing (EMNLP)},
  pages={292--305},
  year={2023}
}

@inproceedings{rohrbach2018object,
  title={Object Hallucination in Image Captioning},
  author={Rohrbach, Anna and Hendricks, Lisa Anne and Burns, Kaylee and Darrell, Trevor and Saenko, Kate},
  booktitle={Proceedings of the 2018 Conference on Empirical Methods in Natural Language Processing (EMNLP)},
  pages={4035--4045},
  year={2018}
}

@article{bai2023qwenvl,
  title={Qwen-{VL}: A Versatile Vision-Language Model for Understanding, Localization, Text Reading, and Beyond},
  author={Bai, Jinze and Bai, Shuai and Yang, Shusheng and Wang, Shijie and Tan, Sinan and Wang, Peng and Lin, Junyang and Zhou, Chang and Zhou, Jingren},
  journal={arXiv preprint arXiv:2308.12966},
  year={2023}
}

@article{wang2024qwen2vl,
  title={Qwen2-{VL}: Enhancing vision-language model's perception of the world at any resolution},
  author={Wang, Peng and Bai, Shuai and Tan, Sinan and Wang, Shijie and Fan, Zhihao and Bai, Jinze and Chen, Keqin and Liu, Xuejing and Wang, Jialin and Ge, Wenbin and others},
  journal={arXiv preprint arXiv:2409.12191},
  year={2024}
}

@article{bai2025qwen25vl,
  title={Qwen2.5-{VL} Technical Report},
  author={Bai, Shuai and Chen, Keqin and others},
  journal={arXiv preprint arXiv:2502.13923},
  year={2025}
}

@article{bai2025qwen3vl,
  title={Qwen3-{VL} Technical Report},
  author={Bai, Shuai and Cai, Yuxuan and Chen, Ruizhe and Chen, Keqin and Chen, Xionghui and Cheng, Zesen and Deng, Lianghao and Ding, Wei and Gao, Chang and Ge, Chunjiang and others},
  journal={arXiv preprint arXiv:2511.21631},
  year={2025}
}

@article{chen2024how,
  title={How Far Are We to {GPT}-4{V}? Closing the Gap to Commercial Multimodal Models with Open-Source Suites},
  author={Chen, Zhe and Wang, Weiyun and Tian, Hao and Ye, Shenglong and Gao, Zhangwei and Cui, Erfei and Tong, Wenwen and Hu, Kongzhi and Luo, Jiapeng and Ma, Zheng and others},
  journal={Science China Information Sciences},
  volume={67},
  number={12},
  pages={220101},
  year={2024},
  publisher={Springer}
}

@article{an2025llavaonevision15,
  title={{LL}a{VA}-{O}ne{V}ision-1.5: Fully Open Framework for Democratized Multimodal Training},
  author={An, Xiang and Xie, Yin and Yang, Kaicheng and Zhang, Wenkang and Zhao, Xiuwei and Cheng, Zheng and Wang, Yirui and Xu, Songcen and Chen, Changrui and Zhu, Didi and others},
  journal={arXiv preprint arXiv:2509.23661},
  year={2025}
}

@inproceedings{liu2024llava15,
  title={Improved Baselines with Visual Instruction Tuning},
  author={Liu, Haotian and Li, Chunyuan and Li, Yuheng and Lee, Yong Jae},
  booktitle={2024 IEEE/CVF Conference on Computer Vision and Pattern Recognition (CVPR)},
  pages={26286--26296},
  year={2024},
  organization={IEEE}
}

@article{zhu2025internvl3,
	title={Intern{VL}3: Exploring advanced training and test-time recipes for open-source multimodal models},
    author={Zhu, Jinguo and Wang, Weiyun and Chen, Zhe and Liu, Zhaoyang and Ye, Shenglong and Gu, Lixin and Tian, Hao and Duan, Yuchen and Su, Weijie and Shao, Jie and others},
    journal={arXiv preprint arXiv:2504.10479},
    year={2025}
}

@article{zhang2023multimodalcot,
  title={Multimodal Chain-of-Thought Reasoning in Language Models},
  author={Zhang, Zhuosheng and Zhang, Aston and Li, Mu and Zhao, Hai and Karypis, George and Smola, Alex},
  journal={Transactions on Machine Learning Research},
  year={2023},
  url={https://openreview.net/forum?id=y1pPWFVfvR}
}

@article{peng2023kosmos2,
  title={Kosmos-2: Grounding multimodal large language models to the world},
  author={Peng, Zhiliang and Wang, Wenhui and Dong, Li and Hao, Yaru and Huang, Shaohan and Ma, Shuming and Wei, Furu},
  journal={arXiv preprint arXiv:2306.14824},
  year={2023}
}

@inproceedings{you2024ferret,
  title={Ferret: Refer and Ground Anything Anywhere at Any Granularity},
  author={You, Haoxuan and Zhang, Haotian and Gan, Zhe and Du, Xianzhi and Zhang, Bowen and Wang, Zirui and Cao, Liangliang and Chang, Shih-Fu and Yang, Yinfei},
  booktitle={International Conference on Learning Representations (ICLR)},
  year={2024},
  volume={2024},
  pages={57153--57180}
}

@inproceedings{man2025argus,
  title={Argus: Vision-centric reasoning with grounded chain-of-thought},
  author={Man, Yunze and Huang, De-An and Liu, Guilin and Sheng, Shiwei and Liu, Shilong and Gui, Liang-Yan and Kautz, Jan and Wang, Yu-Xiong and Yu, Zhiding},
  booktitle={2025 IEEE/CVF Conference on Computer Vision and Pattern Recognition (CVPR)},
  pages={14268--14280},
  year={2025},
  organization={IEEE}
}

@article{yang2023setofmark,
  title={Set-of-Mark Prompting Unleashes Extraordinary Visual Grounding in {GPT}-4{V}},
  author={Yang, Jianwei and Zhang, Hao and Li, Feng and Zou, Xueyan and Li, Chunyuan and Gao, Jianfeng},
  journal={arXiv preprint arXiv:2310.11441},
  year={2023}
}

@inproceedings{hudson2019gqa,
  title={{GQA}: A New Dataset for Real-World Visual Reasoning and Compositional Question Answering},
  author={Hudson, Drew A and Manning, Christopher D},
  booktitle={2019 IEEE/CVF Conference on Computer Vision and Pattern Recognition (CVPR)},
  pages={6693--6702},
  year={2019},
  organization={IEEE}
}

@inproceedings{chen2025acknowledging,
  title={Acknowledging Focus Ambiguity in Visual Questions},
  author={Chen, Chongyan and Tseng, Yu-Yun and Li, Zhuoheng and Venkatesh, Anush and Gurari, Danna},
  booktitle={2025 IEEE/CVF International Conference on Computer Vision (ICCV)},
  pages={1228--1238},
  year={2025},
  organization={IEEE}
}

@misc{park2026gqa,
	title={{GQA-Q2Q}: A Large-scale Dataset for Resolving Entity Ambiguity in Visual Question-Answering via Clarifying Subquestion},
  author={Park, Gyu-Min and Park, Seong-Bae},
	howpublished={Submitted to ICLR 2026. \url{https://openreview.net/forum?id=309pxoSvWr}},
	year={2026}
}

@inproceedings{masry2022chartqa,
  title={Chart{QA}: A Benchmark for Question Answering about Charts with Visual and Logical Reasoning},
  author={Masry, Ahmed and Long, Do Xuan and Tan, Jia Qing and Joty, Shafiq and Hoque, Enamul},
  booktitle={Findings of the Association for Computational Linguistics: ACL 2022},
  pages={2263--2279},
  year={2022}
}

@inproceedings{methani2020plotqa,
  title={Plotqa: Reasoning over scientific plots},
  author={Methani, Nitesh and Ganguly, Pritha and Khapra, Mitesh M and Kumar, Pratyush},
  booktitle={2020 IEEE Winter Conference on Applications of Computer Vision (WACV)},
  pages={1516--1525},
  year={2020},
  organization={IEEE}
}

@inproceedings{mathew2022infographicvqa,
  title={Infographic{VQA}},
  author={Mathew, Minesh and Bagal, Viraj and Tito, Rub{\`e}n and Karatzas, Dimosthenis and Valveny, Ernest and Jawahar, CV},
booktitle={2022 IEEE/CVF Winter Conference on Applications of Computer Vision (WACV)},
pages={2582--2591},
year={2022},
organization={IEEE}
}

@inproceedings{lu2024mathvista,
  title={Mathvista: Evaluating mathematical reasoning of foundation models in visual contexts},
  author={Lu, Pan and Bansal, Hritik and Xia, Tony and Liu, Jiacheng and Li, Chunyuan and Hajishirzi, Hannaneh and Cheng, Hao and Chang, Kai-Wei and Galley, Michel and Gao, Jianfeng},
  booktitle={International Conference on Learning Representations},
  volume={2024},
  pages={23439--23554},
  year={2024}
}

@inproceedings{liu2024mmbench,
	title={{MMbench}: Is your multi-modal model an all-around player?},
	author={Liu, Yuan and Duan, Haodong and Zhang, Yuanhan and Li, Bo and Zhang, Songyang and Zhao, Wangbo and Yuan, Yike and Wang, Jiaqi and He, Conghui and Liu, Ziwei and others},
	booktitle={European conference on computer vision},
	pages={216--233},
	year={2024},
	organization={Springer}
}

@article{chen2024mmstar,
	title={Are we on the right way for evaluating large vision-language models?},
	author={Chen, Lin and Li, Jinsong and Dong, Xiaoyi and Zhang, Pan and Zang, Yuhang and Chen, Zehui and Duan, Haodong and Wang, Jiaqi and Qiao, Yu and Lin, Dahua and others},
	journal={Advances in Neural Information Processing Systems},
	volume={37},
	pages={27056--27087},
	year={2024}
}

@inproceedings{wu2024vstar,
  title={V*: Guided Visual Search as a Core Mechanism in Multimodal {LLM}s},
  author={Wu, Penghao and Xie, Saining},
  booktitle={Proceedings of the IEEE/CVF Conference on Computer Vision and Pattern Recognition (CVPR)},
  year={2024},
  url={https://arxiv.org/abs/2312.14135}
}

@inproceedings{wang2025divide,
  title={Divide, conquer and combine: A training-free framework for high-resolution image perception in multimodal large language models},
  author={Wang, Wenbin and Ding, Liang and Zeng, Minyan and Zhou, Xiabin and Shen, Li and Luo, Yong and Yu, Wei and Tao, Dacheng},
  booktitle={Proceedings of the AAAI Conference on Artificial Intelligence},
  volume={39},
  pages={7907--7915},
  year={2025}
}

@inproceedings{tong2024cambrian1,
  title={Cambrian-1: A Fully Open, Vision-Centric Exploration of Multimodal {LLM}s},
  author={Tong, Shengbang and Brown, Ellis and Wu, Penghao and Woo, Sanghyun and Middepogu, Manoj and others},
  booktitle={Advances in Neural Information Processing Systems},
  year={2024}
}
\bibliographystyle{iclr2027_conference}

\newpage
\appendix
\section{Appendix}

\subsection{Details on our curated dataset for SFT and GRPO} \label{app:dataset}
In this section, we provide comprehensive details regarding the data generation and filtering pipeline used to train GazeVLM. 
We use four seed datasets: GQA \citep{hudson2019gqa}, ChartQA \citep{masry2022chartqa}, PlotQA \citep{methani2020plotqa}, and InfoVQA \citep{mathew2022infographicvqa}.

To ensure that a sample requires multi-region reasoning, we leverage Qwen3-VL-235B-A22B to filter out questions solvable via one-shot perception. 
Furthermore, we explicitly filter the GQA dataset to remove subjective, ambiguous, or multi-answer questions (e.g., generic ``who'' or ``where'' queries) \citep{chen2025acknowledging, park2026gqa}.

For the remaining rigorous queries, we synthesize 10 candidate reasoning trajectories per sample using Qwen3-VL-235B-A22B as our expert model. To prevent erroneous trajectories from degrading our policy, we apply two strict filtering heuristics:

\begin{itemize}
	\item \textbf{Structural validity}: Firstly, we verify that the traces strictly respect the format that we provided. We filter out the rollouts which do not satisfy this criterion and also the ones which have the incorrect final answer. We follow up to attribute a score related to the amount of \texttt{<LOOK>} operations which are performed in the trajectory. Indeed, we encourage traces where the model looks two or three times in different areas, but we start penalizing traces where there are more than five looks as typically it becomes more verbose unnecessarily. We also encourage traces where the model thinks, meaning where the \texttt{<THINK>} tokens are used. Overall, a score is given to each candidate trace and the traces are sorted with respect to that for the following step.
	
	\item \textbf{Spatial grounding check}: We proceed with verifying the grounding claims of the best scored candidate traces. Since our mechanism is not naturally present in the model, we need to make sure that the model does not hallucinate as it generates \texttt{<LOOK>} tokens with their observations. When the model says it focuses on an area in its trace, it generates bounding box coordinates and what it wants to look at. We ensure that the claims given within \texttt{<LOOK>} are correct, using the predicted bounding box and super-imposing it on the image. We then prompt Qwen3-VL-235B-A22B to verify that the description is contained within the highlighted region. If a single \texttt{<LOOK>} operation contains incorrect claims, we prompt the model to refine the coordinates until satisfaction; otherwise, the candidate is discarded. The first candidate (by score) to pass the grounding check is accepted and the others are discarded.
\end{itemize}

Following this pipeline, we extracted a highly curated subset of 11,080 structurally perfect trajectories for our initial Supervised Fine-Tuning (SFT) phase. Examples of generated traces are provided in \Cref{fig:chartqa_1317_t0}.

\paragraph{GRPO Dataset Preparation.} For our RL phase, we require a balanced distribution of difficulty. Using the SFT-trained model, we generate 8 distinct rollouts for the training set and evaluate the empirical success rate. Following \citet{bae2026online}, we filter out excessively easy (near 100\% success) and impossible (0\% success) samples. This results in a highly calibrated subset of 4,453 samples utilized strictly for GRPO optimization. 

\clearpage

% Auto-generated by visualize_traces.py -- do not edit by hand
% Include with: \input{figures}
% Requires: \usepackage{graphicx,xcolor}

\definecolor{gazeA}{HTML}{3b82f6}
\definecolor{gazeB}{HTML}{22c55e}
\definecolor{gazeC}{HTML}{f59e0b}
\definecolor{gazeD}{HTML}{ef4444}
\definecolor{gazeE}{HTML}{a855f7}
\definecolor{gazeF}{HTML}{06b6d4}
\definecolor{gazeG}{HTML}{f97316}
\definecolor{gazeH}{HTML}{ec4899}

\begin{figure}[t]
  \centering
  \begin{minipage}[t]{0.48\linewidth}
    \vspace{0pt}
    \includegraphics[width=\linewidth]{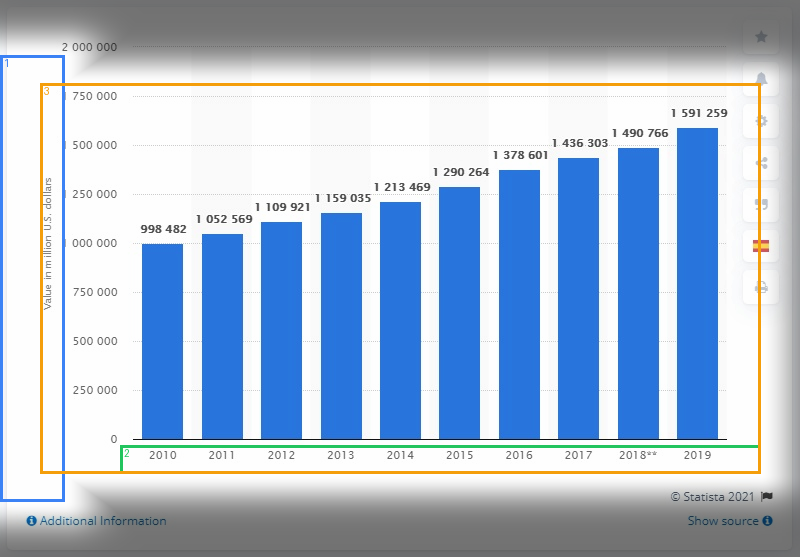}
  \end{minipage}\hfill%
  \begin{minipage}[t]{0.50\linewidth}
    \vspace{0pt}
    \small\raggedright
    \textbf{Q:}~What was the value of property, plant and equipment in service between 2010 and 2019?\\[4pt]
    \textit{\textcolor{gray}{\scriptsize I need to find the value of property, plant and equipment in service for each year from 2010 to 2019 as shown in the bar chart. I will examine the chart's axes to understand the\ldots}}\\[2pt]
    \textcolor{gazeA}{\textbf{1.}~\textit{y-axis label}~{\tiny [0.00, 0.10, 0.08, 0.90]}}\\[-1pt]
    \hspace*{0.7em}{\scriptsize Value in million U.S. dollars}\\[3pt]
    \textcolor{gazeB}{\textbf{2.}~\textit{x-axis labels}~{\tiny [0.15, 0.80, 0.95, 0.85]}}\\[-1pt]
    \hspace*{0.7em}{\scriptsize 2010, 2011, 2012, 2013, 2014, 2015, 2016, 2017, 2018**, 2019}\\[3pt]
    \textcolor{gazeC}{\textbf{3.}~\textit{bar values 2010-2019}~{\tiny [0.05, 0.15, 0.95, 0.85]}}\\[-1pt]
    \hspace*{0.7em}{\scriptsize 2010: 998 482, 2011: 1 052 569, 2012: 1 109 921, 2013: 1 159 035, 2014: 1 213 469, 2015: 1 290 264, 2016: 1 378 601, 2017: 1 436 303, 2018**: 1 490 766, 2019: 1 591 259}\\[3pt]
    \textit{\textcolor{gray}{\scriptsize The chart displays the value of property, plant and equipment in service annually from 2010 to 2019 in million U.S. dollars. I have extracted the specific values for each year. \ldots}}\\[2pt]
    \smallskip\textbf{Answer:}~\texttt{1591259}~\textcolor{gazeB}{$\checkmark$}\enspace{\scriptsize GT:~\texttt{1591259}}
  \end{minipage}
  \vspace{6pt}\par\noindent\rule{\linewidth}{0.4pt}\vspace{6pt}
  \begin{minipage}[t]{0.48\linewidth}
    \vspace{0pt}
    \includegraphics[width=\linewidth]{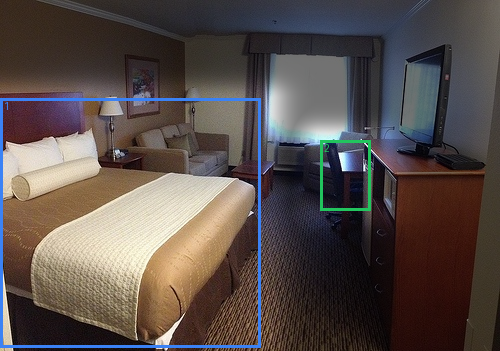}
  \end{minipage}\hfill%
  \begin{minipage}[t]{0.50\linewidth}
    \vspace{0pt}
    \small\raggedright
    \textbf{Q:}~Is the chair to the right or to the left of the bed?\\[4pt]
    \textit{\textcolor{gray}{\scriptsize I need to locate both the bed and the chair in the image to determine their relative positions. I will first identify the bed, then find the chair and compare their horizontal p\ldots}}\\[2pt]
    \textcolor{gazeA}{\textbf{1.}~\textit{bed}~{\tiny [0.00, 0.28, 0.52, 0.99]}}\\[-1pt]
    \hspace*{0.7em}{\scriptsize large bed with brown headboard, white pillows, beige quilted blanket, positioned on left side of room}\\[3pt]
    \textcolor{gazeB}{\textbf{2.}~\textit{chair}~{\tiny [0.64, 0.40, 0.74, 0.60]}}\\[-1pt]
    \hspace*{0.7em}{\scriptsize black office-style chair with wheels, positioned near desk and window}\\[3pt]
    \textit{\textcolor{gray}{\scriptsize The bed is clearly on the left side of the image, occupying the left half of the room. The chair is located further to the right, near the desk and window. Therefore, the chair \ldots}}\\[2pt]
    \smallskip\textbf{Answer:}~\texttt{right}~\textcolor{gazeB}{$\checkmark$}\enspace{\scriptsize GT:~\texttt{right}}
  \end{minipage}
  \caption{What was the value of property, plant and equipment in service between 2010 and 2019? \textit{(top)}; Is the chair to the right or to the left of the bed? \textit{(bottom)}.}
  \label{fig:chartqa_1317_t0}
\end{figure}

\paragraph{Trace Generation and System Prompts.}
To synthesize the SFT reasoning traces, we prompted the expert Qwen3-VL-235B-A22B model with a strict set of guardrails. Crucially, we enforced a ``No Oracle'' rule, ensuring the expert model discovers the answer sequentially through \texttt{<LOOK>}s rather than hallucinating spatial coordinates from a global prior. 

The exact instruction prompt provided to the expert model is as follows:

\begin{promptbox}{Expert Model Prompt (SFT Trace Generation)}
\small\ttfamily\raggedright
You are a visual reasoning agent that examines specific image regions step-by-step to answer a question. You do NOT know the answer in advance — you must search the image to find it. \\
GUARDRAILS: \\
- NO ORACLE: In your first <THINK>, state what you need to find, NOT where things are or what the answer is. You discover locations by looking. \\
- STRUCTURE FIRST: For charts/graphs, <LOOK> at the legend and axes before data points. For documents, read headers before details. For scenes, find the main subject before searching around it. \\
- TIGHT CROPS: Bounding boxes must tightly frame the target. No large boxes covering most of the image. \\
- FACTS IN <LOOK>, LOGIC IN <THINK>: Inside <LOOK>, report only what you see (objects, text, colors, numbers). All reasoning, comparison, and math goes in <THINK>.
\end{promptbox}

\clearpage

During actual SFT and GRPO training (as well as test-time inference), the GazeVLM 4B model is initialized with the following lightweight system prompt to define the syntactic interaction loop:

\begin{promptbox}{GazeVLM System Prompt}
\small\ttfamily\raggedright
You are a visual reasoning assistant that examines specific parts of images to answer questions. \\
Always start by thinking about what you need to examine: <THINK> [plan] </THINK> \\
When you need to examine a specific region, use: <LOOK at="[target]" bbox=[x1, y1, x2, y2]> [Observation] </LOOK> \\
Use <THINK> tags for reasoning between observations, and give your final answer in: <ANSWER> [your answer] </ANSWER>
\end{promptbox}

\subsection{Mathematical Breakdown of the GRPO Reward Formulation} \label{app:grpo_details}

As introduced in \Cref{sec:training}, our GRPO reward function balances positive reinforcement for correctness and grounding against strict penalties for inefficiency. Here, we define the exact scalar values for each component of Equation \ref{eq:reward}.

\paragraph{Correctness and grounding.} We want to prevent the model from guessing the right answer without navigating the image. In order to enforce that, we blend together the correctness reward with a visual grounding indicator, $\mathbf{1}_{\textnormal{gaze}}(c)$. This indicator equals 1 if the completion $c$ contains at least one \texttt{<LOOK>} block with a geometrically valid bounding box ($\textnormal{area} \in [0.005, 0.95]$) of the image and a non-empty observation; otherwise, it is 0. Let $\hat{y}$ be the parsed answer obtained from $c$. We define the correctness reward as:
$$
R_{\textnormal{correct}}(c) = \begin{cases} 
	+1.5 & \textnormal{if } \hat{y} \approx y^* \textnormal{ and } \mathbf{1}_{\textnormal{gaze}}(c) = 1 \\ 
	+0.7 & \textnormal{if } \hat{y} \approx y^* \textnormal{ and } \mathbf{1}_{\textnormal{gaze}}(c) = 0 \\ 
	-0.2 & \textnormal{if } \hat{y} \not\approx y^* \textnormal{ and } \mathbf{1}_{\textnormal{gaze}}(c) = 1 \\ 
	-0.5 & \textnormal{if } \hat{y} \not\approx y^* \textnormal{ and } \mathbf{1}_{\textnormal{gaze}}(c) = 0 \\ 
	-1.0 & \textnormal{if no ANSWER}
\end{cases}
$$

\paragraph{Formatting and validity bonuses.} We apply other positive rewards with regard to formatting where $R_{\textnormal{format}}(c) = 0.15$ if the completion strictly contains all the different tokens. We also reward the model with regard to the validity of the predicted bounding boxes during the \texttt{<LOOK>} blocks. Let $N_L$ be the total number of LOOK blocks and $N_{\textnormal{valid}}$ be the subset of geometrically valid bounding box coordinates. We give a proportional validity bonus $R_{\textnormal{bbox}}(c) = 0.10 \times \frac{N_{\textnormal{valid}}}{N_L}$ or $R_{\textnormal{bbox}}(c) = 0$ if we do not have any \texttt{<LOOK>} block.

\paragraph{Efficiency penalties.} A model can very easily start applying many \texttt{<LOOK>} blocks because of how much we reward it to do so. Nevertheless, this can become useless and redundant especially if the bounding boxes start consistently overlapping and are big regions of the image. To mitigate this issue, we penalize redundant gazing by applying an overlap penalty, depending on the mean pairwise IoU across all generated bounding boxes: $R_{\textnormal{overlap}}(c) = -0.15 \times \overline{\mathrm{IoU}}(c)$, with $\overline{\mathrm{IoU}}(c) = \frac{1}{N_L(N_L - 1)} \sum_{i \neq j} \mathrm{IoU}(b_i, b_j)$.
Finally, we also penalize excessive looping with $R_{\textnormal{excess}}(c) = -0.15$ if $N_L \geq 11$, and rambling with a linearly scaling length penalty $R_{\textnormal{len}}(c) = -0.05 \times \frac{\max(0, |c|_{\textnormal{words}} - w_0)}{w_0}$, where $|c|_{\text{words}}$ denotes the word count of completion $c$ and we set $w_0 = 500$ words.

% $$R_{\textnormal{overlap}}(c) = -0.15 \times \overline{\mathrm{IoU}}(c), \quad \text{with} \quad \overline{\mathrm{IoU}}(c) = \frac{1}{N_L(N_L - 1)} \sum_{i \neq j} \mathrm{IoU}(b_i, b_j).$$

\paragraph{GRPO objective.}
During training, we update our policy $\pi_\theta$ by minimizing the following objective:
\begin{equation}
	L(\theta) = - \frac{1}{G} \sum_{i=1}^{G} A_i \log \pi_\theta(c_i | I, x) + \beta\,\mathrm{KL}(\pi_\theta \,\|\, \pi_{\textnormal{ref}}),
	\quad \text{with }
	A_i = \frac{R(c_i) - \mu_G}{\sigma_G + \epsilon},
\end{equation}
where $\pi_{\textnormal{ref}}$ is the frozen reference policy initialized from the SFT model and $\beta$ controls the $\mathrm{KL}$ regularization strength; $A_i$ is the relative advantage, with $\mu_G$ and $\sigma_G$ the mean and standard deviation of rewards $\{R_1, R_2, \ldots, R_G\}$ and $\epsilon = 10^{-8}$.

\subsection{Experimental Details and Evaluation Settings} \label{app:exp_details}

\paragraph{Training specifics and architecture.} 
We train GazeVLM atop the Qwen3-VL-4B-Instruct architecture using Low-Rank Adaptation (LoRA) to maintain computational efficiency. We apply LoRA adapters ($r=64, \alpha=256$, dropout$=0.05$) across all linear projection layers (\texttt{q\_proj, k\_proj, v\_proj, o\_proj, gate\_proj, up\_proj, down\_proj}). 

The initial SFT phase runs for 3 epochs with a learning rate of $1\times10^{-5}$. The subsequent GRPO phase is executed for 1 epoch with a learning rate of $5\times10^{-6}$ and generating 8 rollouts per prompt. During GRPO, the KL divergence penalty ($\beta$) is dynamically scheduled, decaying linearly from $0.04$ to $0.01$ over the first 50\% of training steps to allow the policy to safely drift from the SFT prior as the correctness signal matures.

For our \texttt{<LOOK>} operations, we lock the suppression strength at $\alpha_s=4$ and the spatial plateau falloff at $\sigma = 0.25$. 
All our experiments are run on 8 $\times$ A100-80GB GPUs.

\paragraph{Evaluation settings.} We evaluated our model and other baselines on multiple visual reasoning benchmarks; from general reasoning, to high-resolution reasoning, chart-image reasoning, hallucination, and others. More specifically, we report our main results on MathVista \citep{lu2024mathvista}, MMBench \citep{liu2024mmbench}, MMStar \citep{chen2024mmstar}, CV-Bench \citep{tong2024cambrian1}, ChartQA \citep{masry2022chartqa}, and HRBench \citep{wang2025divide}. Success in these benchmarks demonstrates the robustness of a model on a span of multiple image reasoning problems. We use an external model \texttt{gpt-oss-120b} to judge the answers provided by our model and other baselines and report the accuracy on the different benchmarks.

We compare our results against a wide range of models. These include closed-source models such as GPT-4o, Gemini2.5-Flash-Lite; vanilla open-source VLMs: Qwen2.5-VL \citep{bai2025qwen25vl}, Qwen3-VL \citep{bai2025qwen3vl}, and Qwen3.5-4B. We also compare against reasoning VLMs which use SFT such as Vision-R1-LlamaV-CI \citep{huang2025visionr1} and SPARC \citep{avogaro2026sparc}. Finally, we include RL-trained VLMs work such as Vision-R1 \citep{huang2025visionr1}, Pixel-Reasoner \citep{Wangetal2025}, DeepEyes \citep{Zhengetal2025}, and Ground-R1 \citep{Caoetal2025}.

\subsection{Additional Ablation Study} \label{app:ablation}

\paragraph{Impact of reinforcement learning (GRPO).} Does the model require RL to internalize active vision, or is Supervised Fine-Tuning (SFT) sufficient? In \Cref{tab:ablation_gaze_importance}, we compare the SFT-only variant of our model trained with the gaze bias against the fully optimized GRPO policy. The results indicate a substantial performance margin in favor of GRPO across all benchmarks (e.g., 4.9 points on MathVista). While SFT successfully imparts the syntactic formatting of the LOOK tokens via behavioral cloning, it is insufficient for complex reasoning. The policy optimization provided by GRPO is essential for teaching the model autonomous visual navigation. By utilizing our bespoke reward function in \Cref{eq:reward} to penalize redundant gazing and reward geometrically valid grounding, GRPO forces the model to transition from merely mimicking instructional traces to executing highly efficient, goal-directed foveal fixations.

\paragraph{Judge validation.} In our work, we score answers with \texttt{gpt-oss-120b} under greedy decoding rather than string matching, because model outputs often combine the selected option with its text (e.g.\ ``B. Iceland'' against the reference ``B''), which exact matching after parsing marks as incorrect. To verify that the judge does not change our conclusions, we rescored all HRBench predictions of the base model and GazeVLM with string matching and compared per-sample decisions (Table~\ref{tab:judge}). The two scorers agree on 99.8\% of samples, and the gain of GazeVLM over the base model is almost the same under both (+4.12 with the judge, +4.25 with string matching, averaged over both splits). The few disagreements are formatting cases of the kind above. 
 
\begin{table}[h]
\centering
\caption{Judge (\texttt{gpt-oss-120b}) vs.\ string matching on all
HRBench predictions.}
\label{tab:judge}
\begin{tabular}{llcccc}
\toprule
\textbf{Split} & \textbf{Method} & \textbf{$n$} & \textbf{Judge acc.} & \textbf{String acc.} & \textbf{Agreement (\%)} \\
\midrule
\multirow{3}{*}{HRBench-4k}
  & Qwen3-VL-4B & 800 & 79.50 & 79.38 & 99.88 \\[2pt]
  & GazeVLM     & 800 & 83.38 & 83.25 & 99.88 \\[2pt]
  & $\Delta$    &     & +3.88 & +3.88 &       \\
\midrule
\multirow{3}{*}{HRBench-8k}
  & Qwen3-VL-4B & 800 & 73.62 & 73.38 & 99.75 \\[2pt]
  & GazeVLM     & 800 & 78.00 & 78.00 & 99.75 \\[2pt]
  & $\Delta$    &     & +4.38 & +4.62 &       \\
\midrule
Overall & $\Delta$ & 1{,}600 & +4.12 & +4.25 & 99.8 \\
\bottomrule
\end{tabular}
\end{table}

% \clearpage

\subsection{Qualitative Results}

\begin{figure}[t]
	\centering
    \vspace{1cm}

	\begin{subfigure}{0.32\linewidth}\centering
		\includegraphics[width=\linewidth]{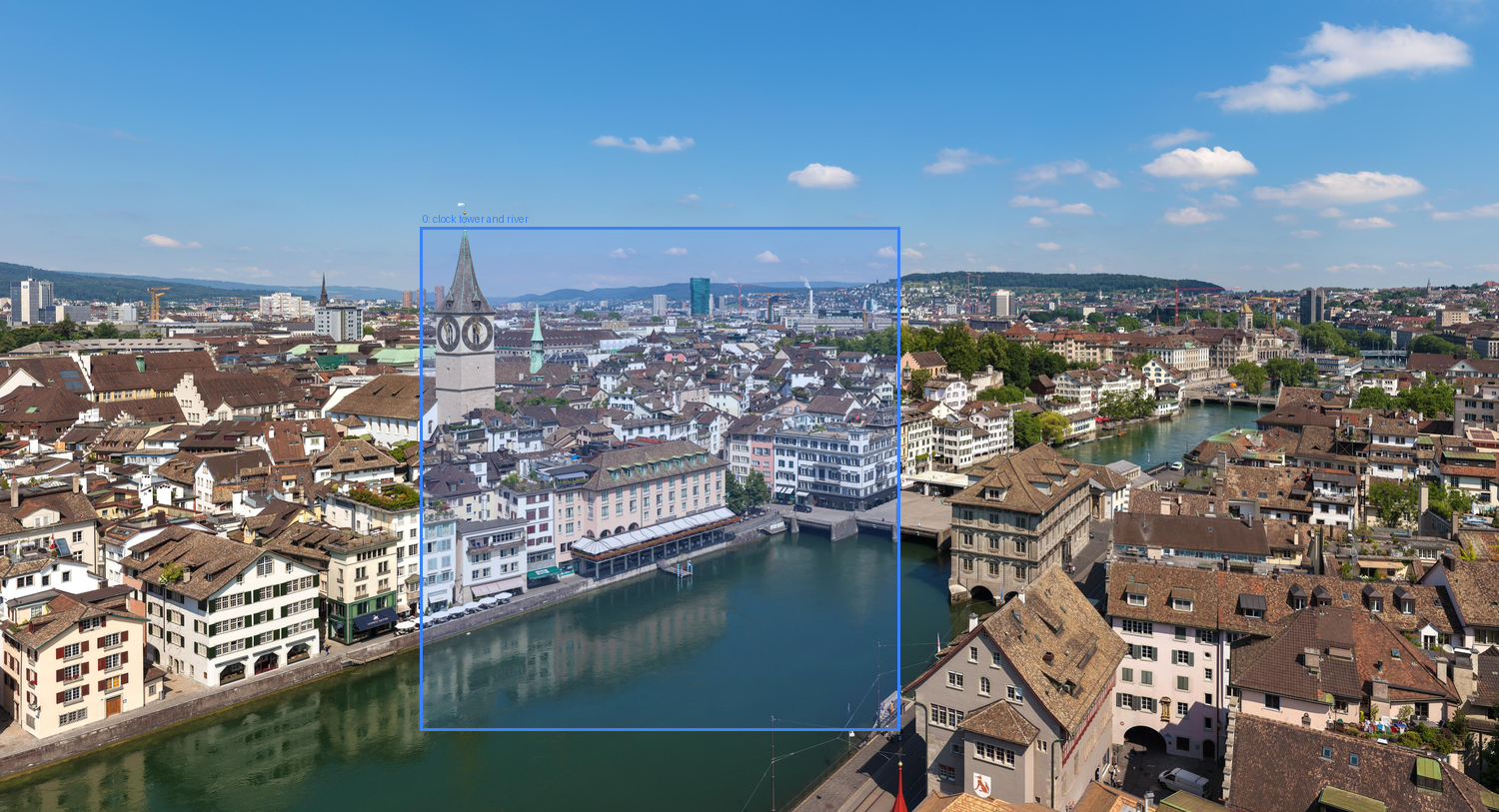}
		\caption{Input with emitted bounding box}
	\end{subfigure}\hfill
	\begin{subfigure}{0.32\linewidth}\centering
		\includegraphics[width=\linewidth, trim=0 0 0 37px, clip]{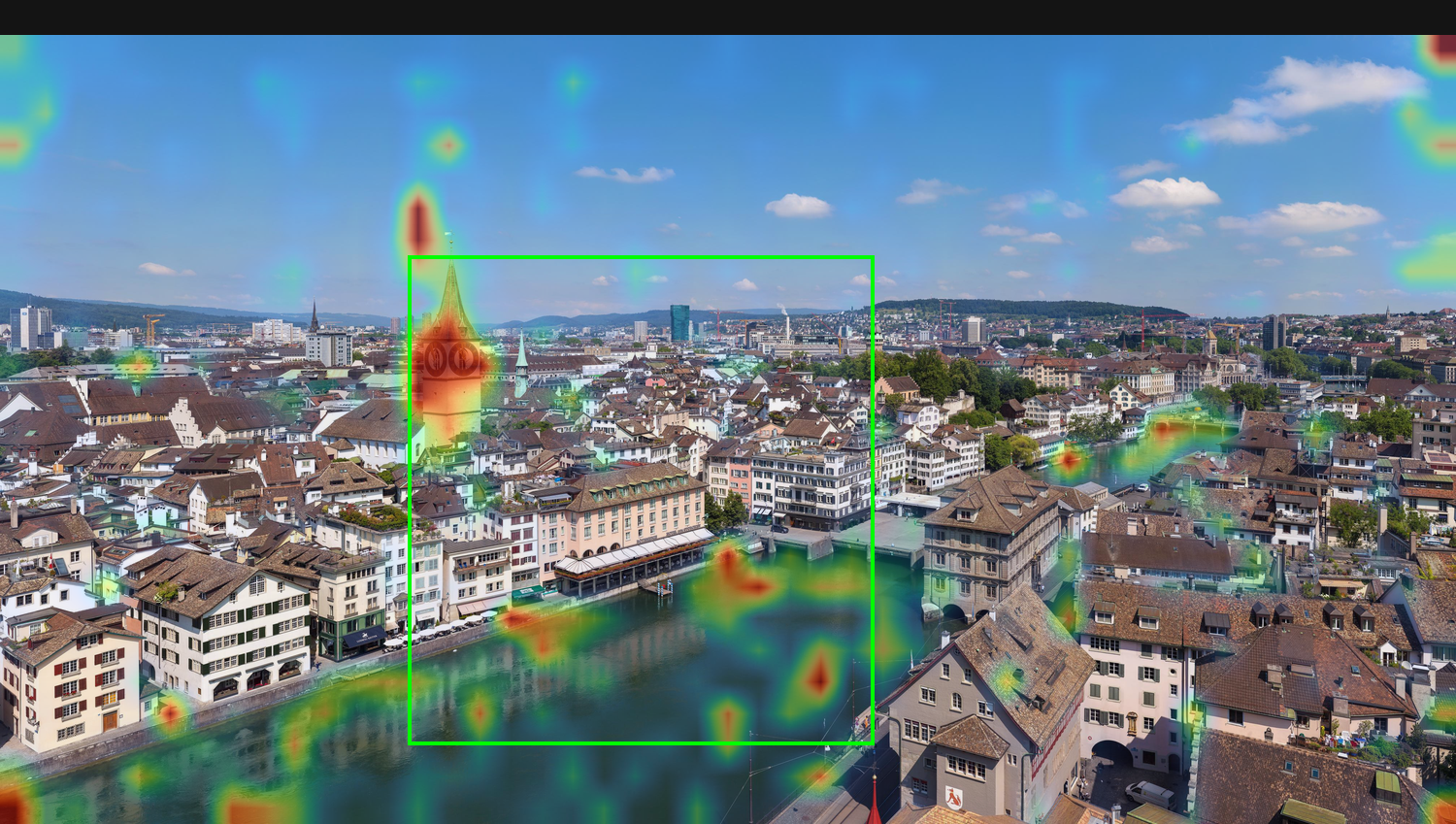}
		\caption{Attention with bias \textsc{off}}
	\end{subfigure}\hfill
	\begin{subfigure}{0.32\linewidth}\centering
		\includegraphics[width=\linewidth, trim=0 0 0 37px, clip]{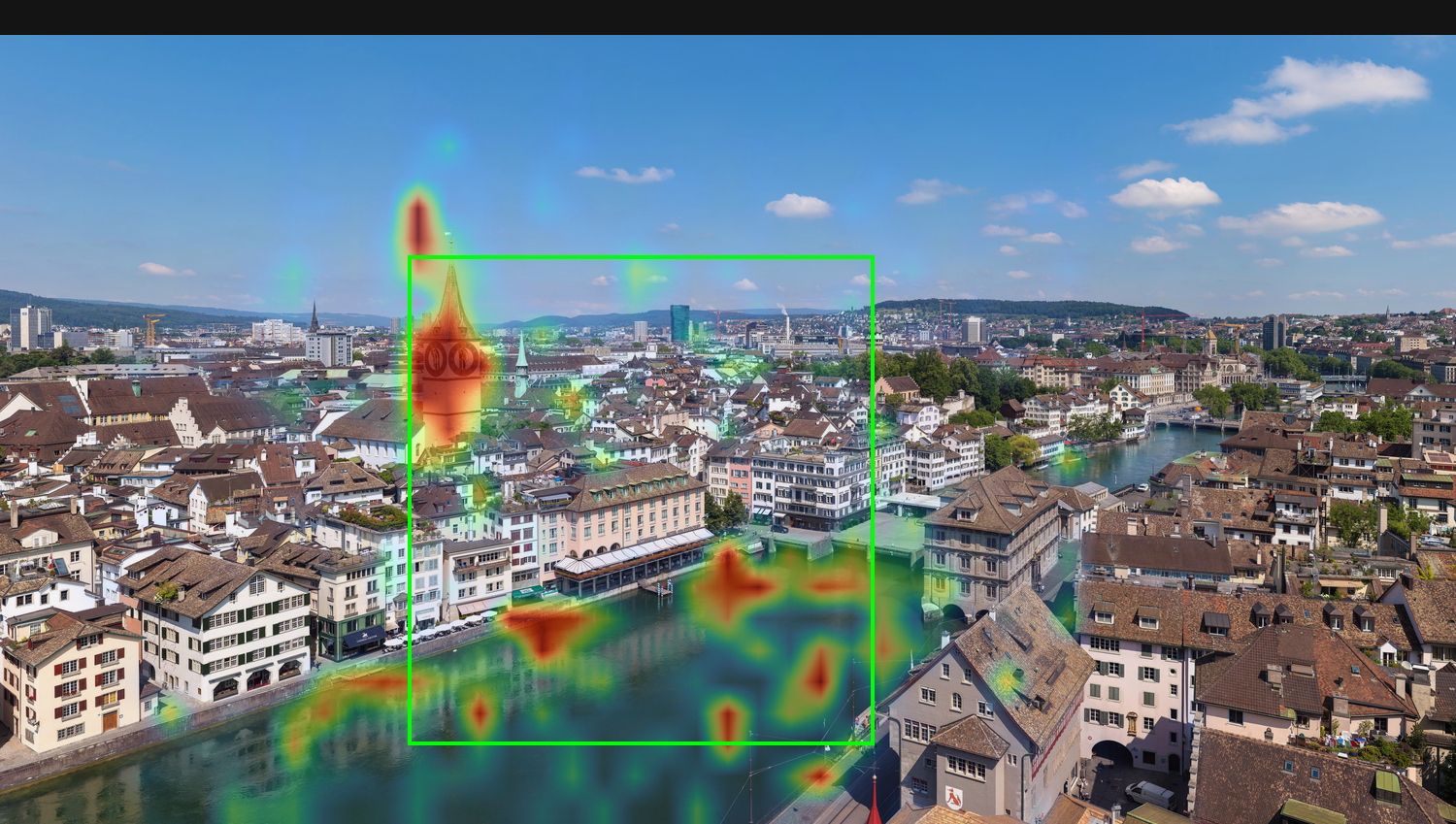}
		\caption{Attention with bias \textsc{on}}
	\end{subfigure}
	\caption{%
		\textbf{Effect of the learned gaze bias on decoder attention.}
		We consider GazeVLM on sample 120 from HRBench-4k, with question ``\emph{Which direction is the river flowing relative to the clock tower?}''. The model is decoded greedily once with the bias inactive (standard GazeVLM inference); the resulting trace is then re-fed through two analysis forward passes over the same prompt and trace, differing only in whether the gaze bias is activated at inference.
		\emph{(a)} the input image with the bounding box the model emitted in its own
		\texttt{<LOOK>} block (in green).
		\emph{(b)-(c)} decoder attention to visual tokens with the bias
		disabled (\textsc{off}) and applied to the emitted bounding box (\textsc{on}).
		Heatmaps show the mean attention from every generated trace token to
		visual tokens, averaged across decoder layers and heads, and share a
		single $5/99$ percentile color scale. The generated trace and decoded
		answer are identical across \emph{(b)} and \emph{(c)}; only the
		analysis forward differs. This isolates the attention bias as the
		proximate cause of the model's focus, independent of any change in
		the output sequence.}
	\label{fig:gaze-mechanism}
\end{figure}

To better understand the mechanistic effect of our learned gaze bias on the model's internal attention, we provide a controlled qualitative analysis in \Cref{fig:gaze-mechanism} on sample 120 from HRBench-4k. Our goal is to isolate the direct effect of the suppression bias on decoder attention, independently of any downstream change in the generated trace or final answer.

To this end, we adopt a two-pass analysis protocol. First, we decode GazeVLM greedily allowing the model to autonomously emit its reasoning trace, including the \texttt{<LOOK>} block with its self-predicted bounding box (shown in green in \Cref{fig:gaze-mechanism}(a)). We then re-feed this exact prompt and trace through two additional forward passes that differ \emph{only} in whether the gaze suppression bias is applied to the emitted bounding box at inference. \Cref{fig:gaze-mechanism}(b) shows the decoder attention with the bias \textsc{off} (standard GazeVLM), while \Cref{fig:gaze-mechanism}(c) shows the same attention with the bias \textsc{on} at inference. Both heatmaps display the mean attention from every generated trace token to the visual tokens, averaged across decoder layers and heads.

Because the trace and decoded answer are identical across (b) and (c), the difference between the two heatmaps should originate from the bias itself. As shown in \Cref{fig:gaze-mechanism}(c), activating the bias visibly concentrates attention onto the targeted region, demonstrating that our suppression mechanism functions as intended at the attention level: it tightly steers the model's foveal focus toward the region the policy has independently identified as task-relevant, while smoothly dampening peripheral visual tokens. This confirms that the gaze bias acts as a clean intervention on attention.

\newcommand{\figpath}{figures/attn_comp/hrbench_4k_120}

\subsection{AI use statement}

In this work, we used generative AI tools for helping with code editing, data generation, evaluation, research, and to improve grammar and readability of the manuscript.
All AI-assisted outputs and edits were reviewed and tested by the authors.

\end{document}